\pdfoutput=1
\documentclass[11pt]{article}
\usepackage[]{acl} % review

\usepackage{times}
\usepackage{latexsym}
\usepackage[T1]{fontenc}
\usepackage[utf8]{inputenc}
\usepackage{microtype}
\usepackage{inconsolata}
\usepackage{graphicx,enumitem,amsmath,amssymb,tikz,xcolor,colortbl,arydshln,subfig,arydshln,booktabs,multirow}

\definecolor{softblue}{RGB}{142,201,253}
\definecolor{softorange}{RGB}{252,180,117}
\definecolor{satblue}{RGB}{40,122,204}
\definecolor{satorange}{RGB}{255,153,51}
\definecolor{satred}{RGB}{255,51,51}

\usepackage{xspace}
\usepackage{mdwlist}
\usepackage{tabularx}
\usepackage{booktabs}
\usepackage{colortbl}

\newcommand{\subrparagraph}[1]{\vspace{0.0mm}\noindent\textit{#1.}}
\newcommand{\rparagraph}[1]{\vspace{1.4mm}\noindent\textbf{#1.}}
\newcommand{\rparagraphnodot}[1]{\vspace{1.2mm}\noindent\textbf{#1}}
\newcommand{\sparagraph}[1]{\vspace{0.0mm}\noindent\textbf{#1.}}
\newcommand{\sparagraphnodot}[1]{\vspace{0.0mm}\noindent\textbf{#1}}

\newcommand*{\inlinecircled}[2][]{\tikz[baseline=(C.base)]{
    \node[inner sep=0pt] (C) {\vphantom{1g}#2};
    \node[draw, circle, inner sep=1pt, yshift=0pt] 
        at (C.center) {\vphantom{1}};}}

\newcommand{\flare}{\texttt{FLARE}\xspace}
\newcommand{\flaremt}{\texttt{FLARE MT}\xspace}
\newcommand{\xmixup}{X-Mixup\xspace}
\newcommand{\inputfusion}{input-level fusion\xspace}

\newcolumntype{Y}{>{\centering\arraybackslash}X}
\definecolor{mintcream}{rgb}{0.96, 1.0, 0.98}
\definecolor{Gray}{gray}{0.92}
\definecolor{lavender}{rgb}{0.9, 0.9, 0.98}
\definecolor{lightgold}{rgb}{1,1,0.88}

\title{Language Fusion for Parameter-Efficient Cross-lingual Transfer}

\author{
    Philipp Borchert\textsuperscript{1,2}, Ivan Vulić\textsuperscript{3}, Marie-Francine Moens\textsuperscript{4} Jochen De Weerdt\textsuperscript{1}\\
    \textsuperscript{1}Research Centre for Information Systems Engineering, KU Leuven, Belgium\\
    \textsuperscript{2}IESEG School of Management, France\\
    \textsuperscript{3}Language Technology Lab, University of Cambridge, United Kingdom\\
    \textsuperscript{4}Department of Computer Science, KU Leuven, Belgium\\
}

\begin{document}
\maketitle
\begin{abstract}
Limited availability of multilingual text corpora for training language models often leads to poor performance on downstream tasks due to undertrained representation spaces for languages other than English. This `under-representation' has motivated recent cross-lingual transfer methods to leverage the English representation space by e.g. mixing English and `non-English' tokens at the input level or extending model parameters to accommodate new languages. However, these approaches often come at the cost of increased computational complexity. We propose \textbf{F}usion for \textbf{La}nguage \textbf{Re}presentations (\flare) in adapters, a novel method that enhances representation quality and downstream performance for languages other than English while maintaining parameter efficiency. \flare integrates source and target language representations within low-rank (LoRA) adapters using lightweight linear transformations, maintaining parameter efficiency while improving transfer performance.
A series of experiments across representative cross-lingual natural language understanding tasks, including natural language inference, question-answering and sentiment analysis, demonstrate \flare's effectiveness. \flare achieves performance improvements of 4.9\% for Llama 3.1 and 2.2\% for Gemma~2 compared to standard LoRA fine-tuning on question-answering tasks, as measured by the exact match metric.\footnote{Our code repository is available at \url{https://github.com/pnborchert/FLARE}}
\end{abstract}

\section{Introduction}

Representation degradation for `non-English' languages poses a challenge in the context of pretrained multilingual language models (mPLMs)\footnote{The domination of the English representation space is observed independent of model architectures, including encoder-only, decoder-only and encoder-decoder transformer~\citep{wu-dredze-2020-languages,lee-etal-2022-pre,yang-etal-2022-enhancing,wendler2024llamas,tang-etal-2024-language}.}. Large-scale English text corpora are widely available for self-supervised pretraining, resulting in superior representation quality and downstream task performance when compared to low(er)-resource languages \citep{lauscher-etal-2020-zero,yang-etal-2022-enhancing}. Despite the substantial improvements, the imbalance in pretraining resources still substantially reduces performance~\citep{winata-etal-2022-cross}.

Cross-lingual transfer (termed XLT henceforth) aims to narrow this performance gap by transferring task-specific knowledge acquired in high-resource languages to lower-resource languages \citep{ruder_2019_survey}. Given the dominance of English in pretraining corpora, machine translations (MT) are frequently utilized to avoid processing non-English data \citep{shi-etal-2010-cross,artetxe-etal-2020-translation,artetxe-etal-2023-revisiting,ansell-etal-2023-unifying}. However, translation can result in information loss, including the loss of cultural nuances, which can negatively impact downstream task performance \citep{conia-etal-2024-towards}. Various XLT techniques address this issue by leveraging both source and target language representation spaces, such as language mixup \citep{yang-etal-2022-enhancing} and concatenating multilingual input sequences for in-context XLT \citep{kim2024crosslingual,tanwar-etal-2023-multilingual,villacueva2024adaptive}. These approaches, while improving XLT, typically focus on representations in a specific mPLM layer or require extensive training and computational resources by extending the input length. 

\begin{figure}
    \centering
    \includegraphics[width=\linewidth]{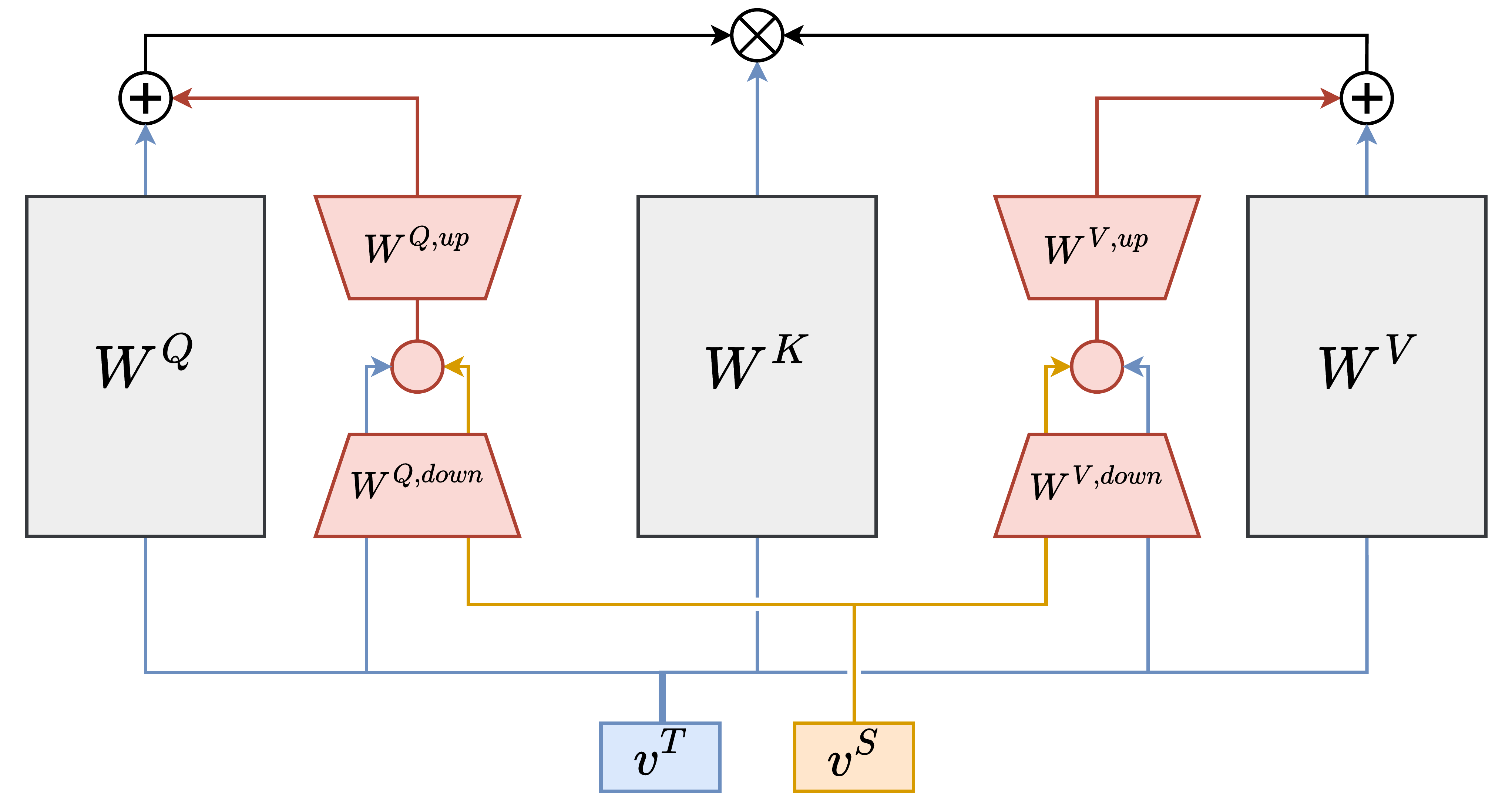}
    \caption{Fusion of \textcolor{satorange}{source} and \textcolor{satblue}{target} representations in LoRA adapters inserted within the query and value matrices. The representations are \textcolor{satred}{fused} in the adapter bottlenecks and the outputs are added \inlinecircled{+} to the query and value outputs before softmax $\otimes$ activation.}
    \label{fig:fusion_lora}
    \vspace{-2mm}
\end{figure}

Parameter-efficient fine-tuning (PEFT) methods are designed to acquire new knowledge and specialize general-purpose models for specific tasks or domains while minimizing the number of extra parameters required and keeping the large underlying mPLM frozen \citep{hu2022lora,pfeiffer2023modular}. In particular, bottleneck-style adapters, such as low-rank adapters (LoRA), extract relevant features from new data by compressing model representations with the assumption that task information can be captured in a lower-dimensional space \citep{pmlr-v97-houlsby19a,hu2022lora}. This directly aligns with the XLT objectives, providing resource-efficient language and task adaptation capabilities. In XLT, adapters are widely used for acquiring task and language knowledge \citep{pfeiffer-etal-2020-mad}. Yet, the extent of knowledge transfer across languages within adapters remains underexplored.

In this work, we introduce \textbf{F}usion for \textbf{La}nguage \textbf{Re}presentations (\flare), a novel approach that merges latent representations from different languages \textit{within lower-dimensional adapter bottlenecks} to enable parameter-efficient XLT. By merging representations from high-resource languages like English into target language representations through lightweight fusion functions, such as addition or multiplication, \flare facilitates effective cross-lingual information transfer with minimal computational overhead. As illustrated in the schematic overview in Figure~\ref{fig:fusion_lora}, \flare performs token-wise fusion of source and target language representations within each transformer block, without adding additional parameters to LoRA and maintaining computational efficiency. \flare is applied during task-specific fine-tuning, enabling models to extract knowledge from both the source and target languages and apply it effectively in the target language.

Our experiments demonstrate \flare's versatility and effectiveness across tasks like natural language inference, sentiment classification, and question answering, using encoder-only, encoder-decoder, and decoder-only multilingual pre-trained language models (mPLMs). It is particularly beneficial for downstream tasks that involve text generation, such as question answering. For instance, \flare improves the absolute exact match performance for Llama 3.1 and Gemma 2 on the TyDiQA dataset by 4.9\% and 2.2\%, respectively. Further experiments illustrate that computational efficiency can be further enhanced by using \textit{latent translations} as source language inputs in \flare, and demonstrate the flexibility of the method, which is orthogonal to the choice of mPLMs and MT systems.

\rparagraph{Contributions}

    \textbf{1)} We introduce \flare, a novel method that fuses language representations within adapter bottlenecks for parameter-efficient cross-lingual transfer.
    \textbf{2)} Our approach improves performance across diverse multilingual downstream tasks, particularly benefiting text generation tasks (e.g., applied to question answering).
    \textbf{3)} We demonstrate the adaptability of our approach by incorporating machine translation encoder representations directly into the mPLM.

\section{Related Work}

\sparagraph{Cross-lingual Representation Transfer} Improving performance for underrepresented languages with mPLMs often involves aligning and combining latent representations from different languages \citep{oh-etal-2022-synergy}. Several methods have been proposed to achieve this, including concatenating multilingual input sequences to leverage a shared representation space \citep{kim2024crosslingual,tanwar-etal-2023-multilingual,villacueva2024adaptive}. Another line of work focuses on projection-based methods, where target language representations are projected onto high-resource languages, such as English, to enhance feature extraction \citep{xu-etal-2023-language-representation}. \citet{yang-etal-2022-enhancing} introduced \xmixup, which combines source and target representations in one specific mPLM layer using cross-attention during downstream task adaptation. Building on this idea, \citet{cao-etal-2023-sharing} proposed using cross-attention with additional semantic and token-level alignment loss terms. In contrast, our \flare method provides a more parameter-efficient approach by directly merging latent source and target language representations within adapter bottlenecks, thereby contributing to the stream of \textit{parameter-efficient XLT}.

Representation fusion has also been applied to integrate information across different modalities, such as vision and language ~\citep{Fang_Wang_Gan_Sun_Liu_2021, ramnath-etal-2021-worldly}. For instance, \citet{qu2024introducing} used feature routing in cross-modal vision-language tasks, guiding language model representations through LoRA bottlenecks using the last hidden state of a vision model. Our work differs in its scope and fusion methodology: \flare extracts richer representations from the source and target languages by capturing layer-wise representations for each transformer block in the mPLMs. Moreover, by ensuring dimensional alignment, we perform token-wise representation fusion within adapter bottlenecks, thereby transferring finer-grained information across languages.

\rparagraph{PEFT in Multilingual Language Models and Cross-Lingual Transfer}
PEFT aims to incorporate task or language-specific knowledge into mPLMs without updating all model weights \citep{pfeiffer-etal-2020-mad}. Most prominent techniques include sparse fine-tuning, which selectively updates model parameters \citep{ansell-etal-2022-composable}, and inserting adapter modules that reduce trainable parameters to a small fraction of total weights of the underlying mPLM~\citep{pmlr-v97-houlsby19a}. Furthermore, PEFT modules are composable, allowing for the combination of information from multiple modules \citep{wang-etal-2022-adamix,lee-etal-2022-fad}. Bottleneck adapters, such as LoRA \citep{hu2022lora} and its variants~\citep{liu2024dora}, are widely used for fine-tuning language models. These adapters project model representations into a lower-dimensional space, creating a bottleneck that regulates information flow \citep{pmlr-v97-houlsby19a}. 

In XLT, mixtures of task and language adapters are used to merge language representation spaces effectively~\citep{lee-etal-2022-fad}. For example, AdaMergeX combines the weights of adapters trained on task data in English with those trained on target language data.  \citep{zhao2024adamergex}. In contrast, our fusion approach enables the model to learn from the interactions between source and target languages by combining inputs from multiple languages during the adapter fine-tuning process. This allows for a more dynamic transfer of task-specific knowledge between languages.

\section{Methodology}

\subsection{Language Representation Fusion}
\label{sec:fusion1}

Our methodology is based on the hypothesis that incorporating English with target language representations enhances cross-lingual knowledge transfer and distills task-relevant information into the target language. We assume (typically MT-created) parallel corpora $\mathcal{P}=\{\left(x^S,x^T\right)\}$ during task fine-tuning, where $x$ are instances in the respective source and target language. Our methodology focuses on using machine-translated parallel data, where widely available English data is translated into the target language for training, and target language data is translated into English during inference. This approach reflects practical real-world scenarios, as human-annotated data is often scarce and costly.

\citet{yang-etal-2022-enhancing} introduced cross-lingual manifold mixup (\xmixup), aligning multilingual representations within a specific transformer layer using consistency loss terms and a cross-attention module. However, this method introduces additional model parameters and shows performance variability depending on the choice of the mixup layer.
Another potentially effective method for aligning multilingual representations is to concatenate source and target language input sequences $x^{S,T}=[x^S;x^T]$ where $x \in \mathbb{R}^{2m}$, with $m$ representing the sequence length of both source and target languages. This so-called \emph{\inputfusion} enables cross-lingual knowledge transfer across all layers of the mPLM, facilitating in-context learning, which typically does not require additional training \citep{villacueva2024adaptive}. However, this approach can become computationally expensive, especially for longer inputs, due to increased input sequence lengths and encounters scalability issues related to the context length limitations in mPLMs.

\begin{figure}
    \centering
    \includegraphics[width=0.8\linewidth]{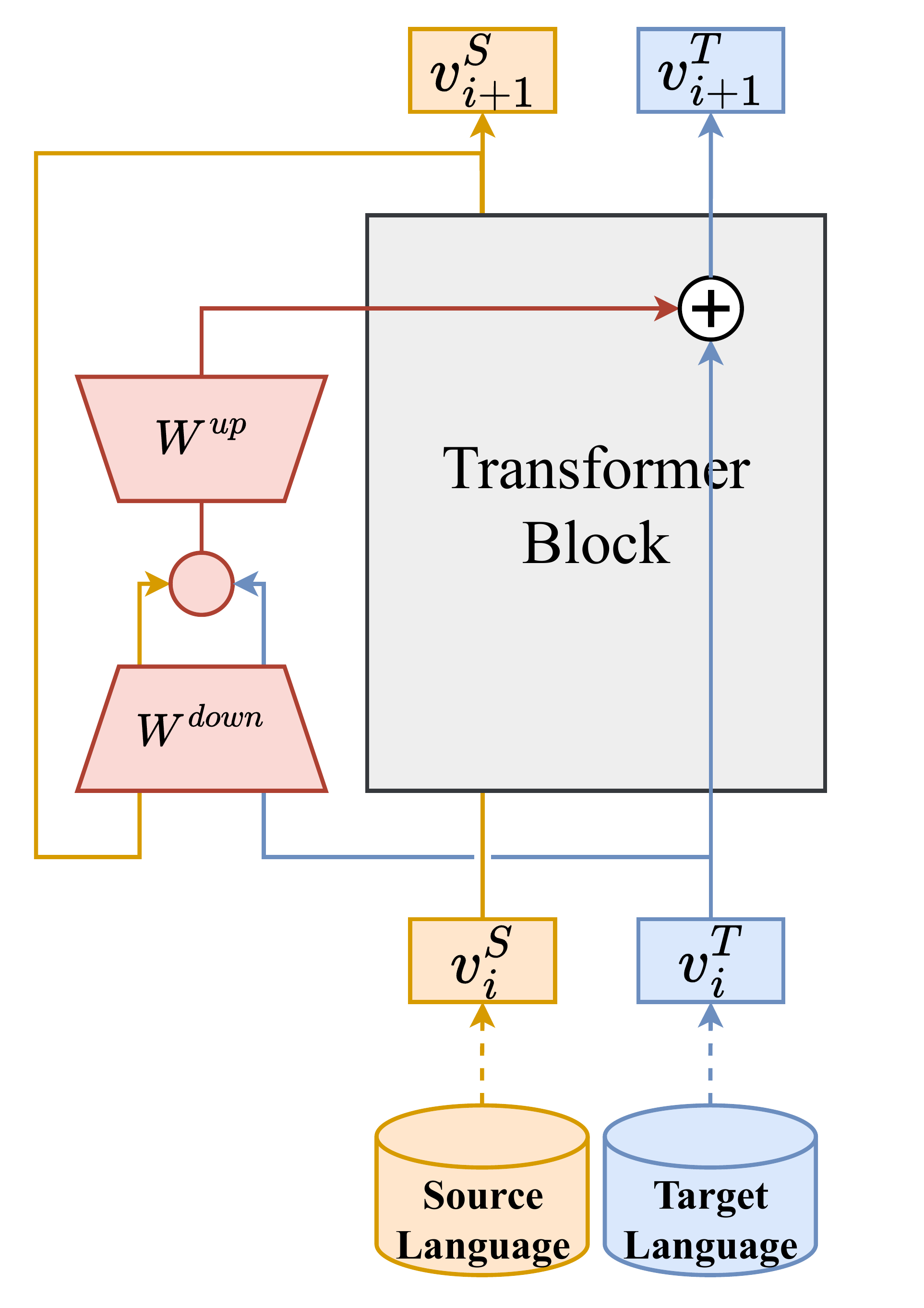}
    \caption{During the forward pass with \flare, \textcolor{satorange}{source} language representations $x^{S}$ are processed by transformer block $i$ and before fusion with \textcolor{satblue}{target} language representations $x^{T}$. Source representations are obtained by inferencing the mPLM without the fusion adapters.}
    \label{fig:forward}
    \vspace{-2mm}
\end{figure}

To address these limitations, we propose \flare, a method for \textit{representation-level} \textit{language fusion} within bottleneck adapters, as illustrated in Figure~\ref{fig:forward}. Instead of extending the input, \flare processes source and target language representations independently and fuses them only within the adapters, thus preserving computational efficiency. Source language representations $v_i^S$, extracted from the frozen mPLM without adapters, and target language representation $v_i^T$ at transformer block $i$ are down-projected using $W^{down}$ and combined with fusion function $\phi$ (see Section \ref{sec:fusion2}) to create a fused representation $h=\phi\left(v_{i+1}^S W^{down},v_i^T W^{down}\right)$, where $h \in \mathbb{R}^{m \times r}$ with sequence length $m$ and bottleneck dimensions $r$ \citep{qu2024introducing}. We utilize the source representation $v_{i+1}^S$, which has been processed by the subsequent transformer block, to leverage task-specific information extracted from the source language. Following a standard LoRA procedure, this fused low-rank representation is then up-projected and added to the frozen attention outputs $v^{0}$ to form the target language output representation $v_{i+1}^{T} = h W^{up} + v^{0}$ of the attention block. 

This enhances model performance during task adaptation in the target language by directing the model's attention to task-relevant information. Thereby, the adapter bottleneck is used for cross-lingual knowledge transfer, as well as task and language adaptation. A key advantage of \flare is the reduction in computational complexity, thereby enhancing parameter efficiency for both task and language adaptation. By processing multilingual inputs separately and only fusing highly compressed representations within adapter bottlenecks, our method avoids the computational overhead associated with quadratic scaling in attention computations for model dimensions $d$, thus enhancing resource efficiency. Furthermore, the memory requirements are limited to the last hidden states obtained from the output of each transformer block.

% Assuming that new task information can be learned within low-rank adapters, we posit that task-specific cross-lingual knowledge can be effectively transferred within adapter bottlenecks. This enhances efficiency, and also compresses and aligns task-relevant information, simplifying the complexity of representations $r \ll d$.
% The down-projection within the bottleneck adapters is applied to both target and source language representations, exploiting the unified embedding space acquired during self-supervised pretraining for cross-lingual adaptation.

Moreover, our fusion approach is agnostic regarding the source language representation. We exploit this flexibility in the \flaremt variant, which explores the impact of reducing computational resources for processing the source language on cross-lingual transfer performance. Specifically, \flaremt utilizes representations from a MT encoder $\mathcal{M}$ as `latent translations' that serve as source language representations. This avoids discretizing the translation as text through the MT decoder. \flaremt further enhances resource efficiency compared to regular \flare by bypassing the forward pass of the source language in the mPLM. We extract a single representation (latent translation) from the MT encoder by processing the target language input $v^T = \mathcal{M}\left(x^T\right)$, where $v^T \in \mathbb{R}^{m \times d_{\mathcal{M}}}$. 

\begin{figure}
    \centering
    \includegraphics[width=0.8\linewidth]{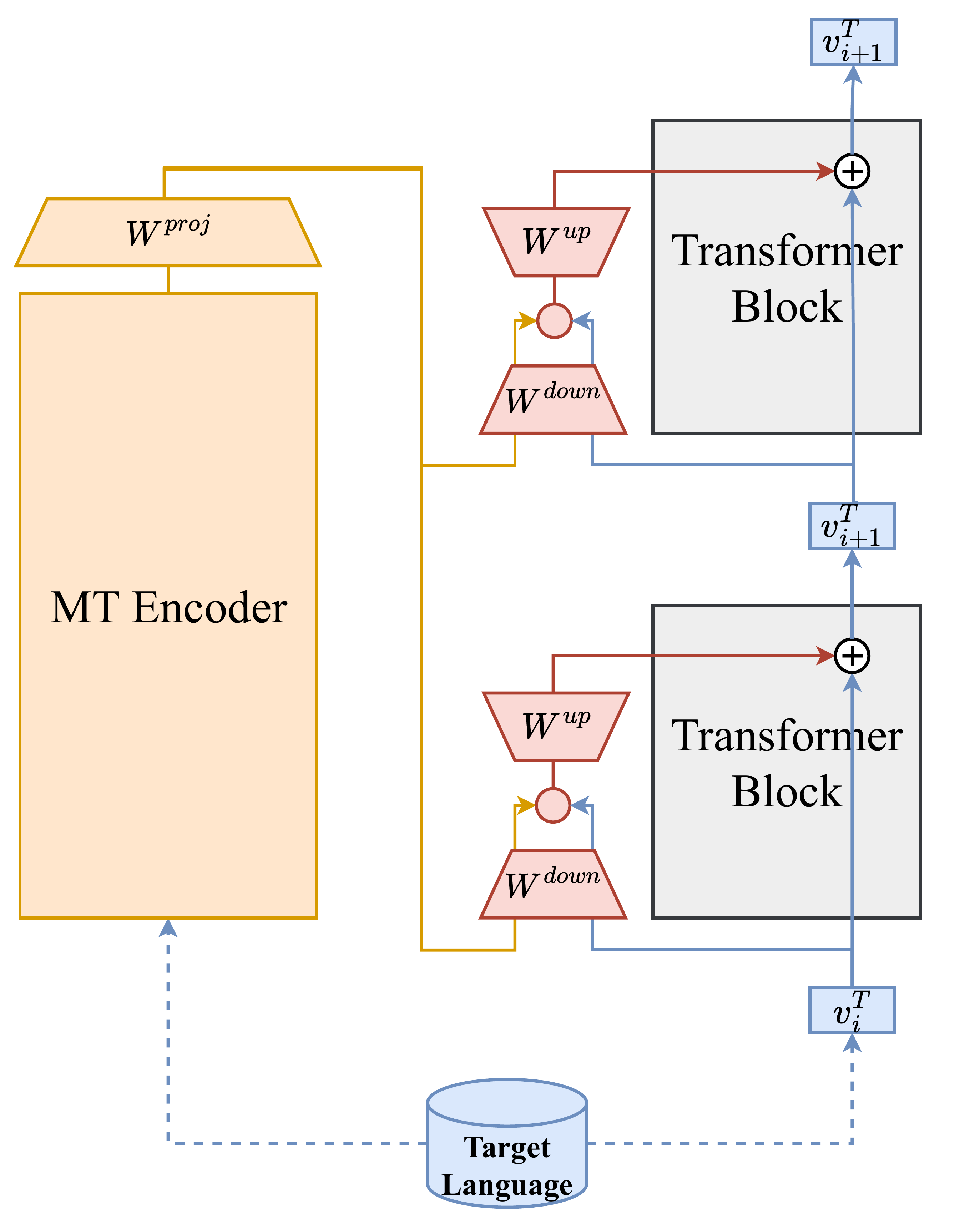}
    \caption{Illustration of the \flaremt variant where projected encoder representations from an MT model are directly fused with target language representations within the fusion adapters in the mPLM. Encoder representations from the MT model serve as latent translations, avoiding discretization in the decoder.}
    \label{fig:fusion_mt}
    \vspace{-2mm}
\end{figure}

To ensure compatibility between the dimensionality of the MT encoder outputs and the mPLM, we utilize a linear projection layer $W^{proj}$. This projection is jointly trained during the adaptation to the downstream task, ensuring resource efficiency. The up-projected representation $v^T W^{proj}$ is fused with the target language representation within the adapter bottlenecks of each mPLM layer, as displayed in Figure~\ref{fig:fusion_mt}.

\subsection{Fusion Functions}
\label{sec:fusion2}
To fuse cross-lingual representations in bottleneck adapters, we evaluate both linear and non-linear transformations that do not require additional model parameters, alongside cross-attention. We extract token-wise representations from source and target language sequences, capturing rich contextual information at the token level.

The down-projected representations in the adapter bottlenecks for source and target languages are denoted as $S=v^{S} W^{down}$ and $T=v^{T} W^{down}$, where $S$ and $T$ are representations of dimensions $\mathbb{R}^{m \times r}$. These representations are subsequently combined at the token level through the following fusion functions:
\begin{enumerate*}[]
    \setlength{\itemsep}{0pt}
    \item element-wise addition (\emph{add}): $S+T$
    % \begin{equation*}
    %     S+T
    % \end{equation*}
    \item element-wise multiplication (\emph{mul}): $S \circ T$
    % \begin{equation*}
    %     S \circ T
    % \end{equation*}
    % \item matrix multiplication (\emph{matmul}): $ST'T$
    % \begin{equation*}
    %     ST'T
    % \end{equation*}
    \item \emph{cross-attention}:\footnote{Although cross-attention modules add parameters to the adapters, the low bottleneck dimensions $r$, typically smaller than 64, minimize the parameter count in comparison to the model's internal dimensions $d$. Specifically, we utilize a single cross-attention head to maintain efficiency.} {\small $\text{softmax}\left(\frac{W^Q_aS\left(W^K_aT\right)'}{\sqrt{r}}\right)W^V_aT$}

\end{enumerate*}
$W^Q_a$, $W^K_a$ and $W^V_a$ are the weight matrices of the query, key and value projections in the adapter $a$, respectively, and $'$ denotes the matrix transpose. We focus on lightweight linear transformations to maintain parameter and computational efficiency.

Additionally, linear fusion functions are extended with non-linear transformations through rectified linear units $ReLU\left(S\right)$ and $ReLU\left(T\right)$ \citep{qu2024introducing}. This allows for selective information flow in token representations, which can be particularly beneficial for multilingual input sequences that may be misaligned at the token level. By introducing non-linear transformation functions, we can restrict the propagation of misaligned information, potentially leading to improved task performance.

\subsection{Training}

To adapt the mPLM to downstream tasks in the target language, we insert LoRA fusion adapters into the query and value weight matrices of the mPLM that has been previously fine-tuned on English task data, referred to as the \textit{base model}. These adapters implement fusion function $\phi$ that combines source and target language input representations into a single fused representation. Consistent with standard PEFT training, only the task head and LoRA parameters are trainable, while all other parameters remain frozen.

During the forward pass, illustrated in Figure~\ref{fig:forward}, we extract representations from both the source and target languages at each transformer block. Source language representations are obtained from the base model without fusion adapters. These layer-wise representations are stacked in matrix $V^S \in \mathbb{R}^{l \times m \times d}$, where $l$ represents the number of layers in the mPLM. Target language representations are obtained during the forward pass through the base model with fusion adapters. 
In our \flare approach, the source and target language representations are compressed to lower dimensions $r \ll d$ using the adapter's down-projection $W^{down}$. The compressed representations are then combined through the fusion function, and decompressed in the up-projection. By sharing the down-projection layers for both source and target language representations before fusion, we hypothesize that the model's reliance on the English representation space is reduced.

\section{Experimental Setup}

\subsection{Underlying Models and Baselines}

\sparagraph{mPLMs}
First, we again note that the design of \flare makes it transformer architecture-agnostic; it is directly applicable to encoder, decoder and encoder-decoder models. Our experiments are thus based on various mPLMs including the encoder-only XLM-R Large (550M) \citep{conneau-etal-2020-unsupervised}, the encoder-decoder mT5-XL (3.7B) \citep{xue-etal-2021-mt5}, the decoder-only Llama 3.1 (8B) \citep{llama3}, and the decoder-only Gemma 2 (9B) \citep{gemmateam2024}.

\rparagraph{Fine-Tuning Setup}
We follow a modular XLT approach where the mPLM is fine-tuned on English task data and subsequently adapted using task data in the target language \citep{zhao-etal-2021-closer}. For decoder-only models like Llama and Gemma, we use a causal language modeling objective for fine-tuning, and the models generate predictions as text accordingly. We employ the QLoRA fine-tuning approach with 4-bit quantization and insert LoRA adapters in all linear layers for Llama and Gemma models \citep{dettmers2023qlora}. Importantly, we apply representation fusion in \flare only in the attention modules, ensuring a consistent experimental setup across different transformer architectures. The LoRA configurations use $r=64$ and $\alpha=128$, where $r$ corresponds to the bottleneck dimensions and $\alpha$ is a scaling factor applied to the low-rank updates before merging them with the frozen attention outputs. Detailed hyperparameter configurations for each model can be found in Table~\ref{tab:params} in the appendix.

\rparagraph{Baselines}
We evaluate \flare against several baselines, including zero-shot cross-lingual transfer, translate-test, as well as translate-train methods such as regular LoRA fine-tuning, \xmixup, and \inputfusion. We focus on baseline methods that fine-tune on task-specific data in the target language, as these provide the most competitive performance. All translate-train models are trained with the same LoRA configurations. Unless otherwise specified, \flare models are trained using the \emph{add+relu} fusion function, with a detailed comparison of fusion functions presented in Table \ref{tab:results_fusion_functions}. Model checkpoints are selected based on validation data that was machine-translated from English to the respective target languages.

\xmixup aligns source and target language representations through cross-attention in one specific transformer layer and further aligns model outputs using consistency loss terms \citep{yang-etal-2022-enhancing}. In contrast, \inputfusion combines source and target language texts directly in the input prompt of the mPLM, doubling the sequence length \citep{kim2024crosslingual,villacueva2024adaptive}.
More details on the baselines below:

\subrparagraph{Zero-Shot XLT} The base model fine-tuned on English task data is directly evaluated on test data in the target languages without further training.

\subrparagraph{Translate-Test} Test sets in each target language are translated into English using NLLB~\citep{nllbteam2022language}. Subsequently, the base model is evaluated on these machine-translated test sets.\footnote{Although monolingual English-only PLMs can process machine-translated text, they fail to outperform multilingual models, particularly when evaluating low-resource languages or culturally sensitive content \citep{ebing-glavas-2024-translate}.}

\subrparagraph{Translate-Train} The base model is fine-tuned on machine-translated task data in the respective target languages. The training data comprises instances translated from English to the target language using NLLB. For fusion methods and \xmixup, we obtain the required `silver' parallel data also through MT (using NLLB). The training set consists of parallel sets of English and MT-ed instances, whereas the validation and test sets consist of parallel target language instances and corresponding machine translations into English. We posit that the assumed absence of gold translations both during training and during inference is the most realistic evaluation of \flare models. 

\subsection{Evaluation Tasks and Datasets}

\sparagraphnodot{XNLI} consists of machine-translated sentence pairs that are translated from English to 15 languages \citep{conneau-etal-2018-xnli}. The task involves determining whether a sentence entails, contradicts, or is neutral to a given premise.

\rparagraphnodot{NusaX} is a human-annotated sentiment classification dataset that spans 11 Indonesian languages, including low-resource languages \citep{winata-etal-2023-nusax}. With 500 labeled instances for each language, the dataset evaluates few-shot adaptation.

\rparagraphnodot{TyDiQA-GoldP} is a human-annotated extractive QA dataset covering 8 languages \citep{clark-etal-2020-tydi}. The task is to extract the answer spans from context passages.

Additional information on evaluation languages and datasets used for source language fine-tuning are available in Table~\ref{tab:tasks_langs} in the appendix.

\subsection{Machine Translations}
We utilize the NLLB 3.3B variant \citep{nllbteam2022language} as the main MT model, employing greedy decoding to obtain translations \citep{artetxe-etal-2023-revisiting}. Additionally, \flaremt utilizes the encoder of the NLLB 600M variant to generate latent translations. To maintain consistency in our experimental setup, we also translate languages that are not directly supported by NLLB. Specifically, Madurese (mad) and Ngaju (nij) are translated using the Indonesian language identifier, as these languages are not supported by NLLB\footnote{We note that Toba Batak (bbc) is unsupported by NLLB and excluded from the evaluation due to translation artifacts resulting in random classification performance.} \citep{winata-etal-2023-nusax}.
For translating extractive QA datasets, we employ EasyProject \citep{chen-etal-2023-frustratingly}, which involves enclosing answer spans within marker tokens prior to translation with NLLB. This method allows us to determine the position of the translated answer spans by locating these marker tokens in the translated text. Instances that fail to retain the marker tokens in the translated output are excluded from evaluation.

\begin{table*}
\begin{center}
\vspace{-0.5mm}
{
\def\arraystretch{0.99}
\fontsize{8.0pt}{7.9pt}\selectfont
%\begin{tabular}{@{\extracolsep{\fill}} lcccc}
\begin{tabularx}{\textwidth}{l YYYYY}
\toprule
\textbf{Model} & \textbf{XNLI} & \textbf{TyDiQA} & \textbf{NusaX} & \textbf{Avg.} & \textbf{Avg. rank} \\
\midrule
\rowcolor{Gray}
\multicolumn{6}{l}{\textit{\textbf{Zero-Shot Cross-Lingual Transfer} (models are trained on English data)}} \\
\midrule
XLM-R Large & $76.95 \pm 0.3$ & $36.31 \pm 2.3$ & $75.26 \pm 1.0$ & $62.84$ & $2.33$ \\
mT5-XL & $77.92 \pm 1.2$ & $45.90 \pm 0.2$ & $74.72 \pm 1.6$ & $66.18$ & $1.67$ \\
Llama 3.1 8B & $77.40 \pm 0.2$ & $\phantom{0}2.36 \pm 0.2$ & $71.74 \pm 2.8$ & $50.50$ & $3.33$ \\
Gemma 2 9B & $80.47 \pm 0.1$ & $\phantom{0}2.46 \pm 0.2$ & $71.61 \pm 3.4$ & $51.51$ & $2.67$ \\
\midrule
\rowcolor{lavender}
\multicolumn{6}{l}{\textit{\textbf{Translate-Test} (test data is translated to English)}} \\
\midrule
% XLM-R Base & $74.78 \pm 0.4$ & $48.76 \pm 0.8 / 36.94 \pm 1.0$ & $75.93 \pm 0.5$ & $64.52$ \\
XLM-R Large & $77.13 \pm 0.2$ & $41.06 \pm 1.6$ & $74.85 \pm 1.0$ & $64.35$ & $2.67$ \\
mT5-XL & $79.03 \pm 0.2$ & $47.92 \pm 0.2$ & $75.77 \pm 0.3$ & $67.57$ & $1.67$ \\
Llama 3.1 8B & $79.08 \pm 0.5$ & $\phantom{0}2.53 \pm 0.4$ & $72.67 \pm 2.4$ & $51.54$ & $2.67$ \\
Gemma 2 9B & $79.99 \pm 0.9$ & $\phantom{0}2.28 \pm 0.2$ & $71.61 \pm 3.4$ & $51.29$ & $3.00$ \\
\midrule
\rowcolor{mintcream}
\multicolumn{6}{l}{\textit{\textbf{Translate-Train} (models are trained on training data translated to the target language)}} \\
\midrule
XLM-R Large w/ LoRA & $80.49 \pm 1.3$ & $40.14 \pm 0.4$ & $77.00 \pm 0.8$ & $65.88$ & $3.33$ \\
w/ \xmixup \citep{yang-etal-2022-enhancing} & $79.47 \pm 0.2$ & $38.24 \pm 3.2$ & $76.37 \pm 2.8$ & $64.69$ & $4.67$ \\
w/ \inputfusion & $77.24 \pm 0.8$ & $40.45 \pm 0.5$ & $78.53 \pm 0.3$ & $65.41$ & $3.00$ \\
w/ \flaremt & $81.60 \pm 0.3$ & $38.88 \pm 1.3$ & $77.18 \pm 0.2$ & $65.89$ & $2.67$ \\
w/ \flare & $80.99 \pm 0.9$ & $40.93 \pm 0.2$ & $79.18 \pm 1.4$ & $\mathbf{67.03}$ & $\mathbf{1.33}$ \\
\cmidrule{1-6}
mT5-XL w/ LoRA & $79.79 \pm 2.1$ & $46.76 \pm 0.7$ & $80.41 \pm 0.2$ & $68.99$ & $3.67$ \\
w/ \xmixup \citep{yang-etal-2022-enhancing} & $79.63 \pm 1.0$ & $48.23 \pm 0.5$ & $78.61 \pm 0.2$ & $68.82$ & $4.00$ \\
w/ \inputfusion & $78.81 \pm 0.2$ & $47.58 \pm 0.2$ & $80.12 \pm 0.2$ & $68.84$ & $4.33$ \\
w/ \flaremt & $80.80 \pm 1.4$ & $48.48 \pm 0.2$ & $81.37 \pm 0.8$ & $70.22$ & $1.67$ \\
w/ \flare & $81.00 \pm 1.2$ & $49.34 \pm 0.3$ & $80.54 \pm 0.2$ & $\mathbf{70.29}$ & $\mathbf{1.33}$ \\
\cmidrule{1-6}
Llama 3.1 8B w/ LoRA & $80.74 \pm 0.4$ & $42.84 \pm 0.7$ & $74.76 \pm 1.4$ & $66.11$ & $3.00$ \\
w/ \xmixup \citep{yang-etal-2022-enhancing} & $80.22 \pm 0.2$ & $17.47 \pm 1.6$ & $75.91 \pm 0.7$ & $57.87$ & $4.00$ \\
w/ \inputfusion & $80.70 \pm 0.5$ & $46.09 \pm 0.9$ & $74.60 \pm 1.6$ & $67.13$ & $3.33$ \\
w/ \flaremt & $80.83 \pm 0.2$ & $38.95 \pm 0.2$ & $74.52 \pm 1.6$ & $64.77$ & $3.67$ \\
w/ \flare & $80.92 \pm 0.2$ & $47.74 \pm 1.2$ & $76.08 \pm 1.1$ & $\mathbf{68.25}$ & $\mathbf{1.00}$ \\
\cmidrule{1-6}
Gemma 2 9B w/ LoRA & $84.89 \pm 0.4$ & $49.93 \pm 0.7$ & $79.37 \pm 1.2$ & $71.40$ & $2.67$ \\
w/ \xmixup \citep{yang-etal-2022-enhancing} & $84.62 \pm 0.5$ & $35.45 \pm 2.0$ & $79.94 \pm 1.2$ & $66.67$ & $3.67$ \\
w/ \inputfusion & $80.53 \pm 0.2$ & $51.29 \pm 0.3$ & $77.98 \pm 1.1$ & $69.93$ & $4.00$ \\
w/ \flaremt & $84.84 \pm 0.3$ & $49.63 \pm 0.9$ & $78.09 \pm 0.9$ & $70.85$ & $3.67$ \\
w/ \flare & $85.01 \pm 0.4$ & $52.14 \pm 0.7$ & $80.86 \pm 0.5$ & $\mathbf{72.67}$ & $\mathbf{1.00}$ \\
\bottomrule
\end{tabularx}
}
\caption{Average performance (with standard deviation) on natural language understanding datasets, computed over 5 random seeds. Metrics used are: Accuracy for XNLI, Exact Match for TyDiQA, and Macro F1 for NusaX. The best-performing results for each XLT model are highlighted in \textbf{bold}.}

\label{tab:results}
\end{center}
\vspace{-4mm}
\end{table*}

\section{Results and Discussion}
\label{sec:results}

\sparagraphnodot{Main Results.} 
The results displayed in Table \ref{tab:results} confirm our hypothesis that task-specific knowledge can be efficiently transferred from English to other languages within adapter bottlenecks. Our proposed approach, \flare, consistently surpasses all baselines across various tasks, demonstrating robust performance and validating the effectiveness of our method. It improves the average performance, averaged across metrics for all tasks, by 2.14\% and 1.27\% for Llama and Gemma, respectively, when compared to standard LoRA fine-tuning. The most substantial performance gains are observed on the TyDiQA dataset, particularly for text generation tasks with decoder-only models. \flare substantially improves performance on this dataset, with the largest gains achieved on Indonesian, Russian, and Swahili. This suggests that latent representation fusion with \flare works best for text generation when the target languages have a similar word order to the source language, in this case, subject-verb-object. However, we also observe substantial performance gains for the Llama model on Telugu, which has a different word order than English, indicating that \flare can still achieve considerable improvements even when the word order differs.

The results on XNLI and NusaX do not exhibit a clear correlation between performance benefits for languages with subject-verb-object word order.
Furthermore, the results on NusaX demonstrate that \flare can consistently provide performance improvements for lower-resourced languages, even when only a few training data is available. This highlights the potential of \flare to support language adaptation in low-resource settings, where data is scarce. Compared to all benchmarked models, \flare provides consistent performance benefits, demonstrating its effectiveness in transferring knowledge from English to other languages, and its potential to improve the performance of downstream tasks in low-resource languages. Beyond performance benefits, FLARE reduces the average training time on TyDiQA by more than 40\% when compared to input-level fusion.

\begin{figure}
    \centering
    \includegraphics[width=\linewidth]{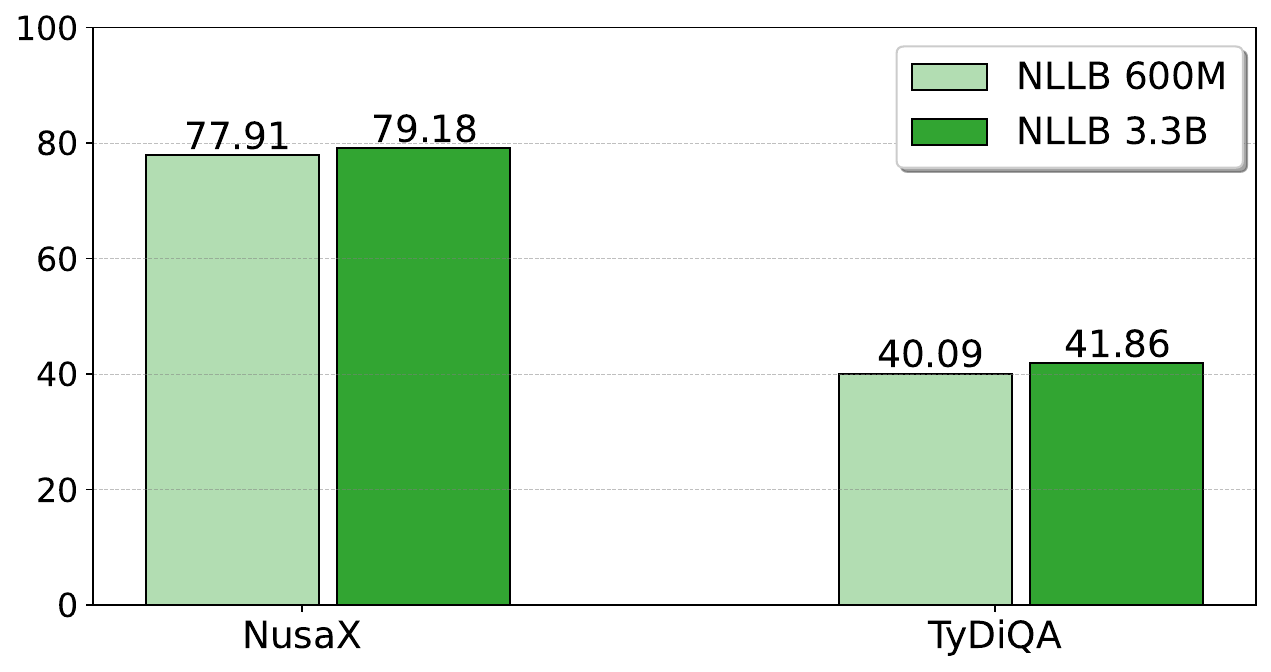}
    \caption{Average performance differences on NusaX and TyDiQA for XLM-R Large using \flare with MT models of different size.}
    \label{fig:mt_models}
    \vspace{-2mm}
\end{figure}

\rparagraph{Impact of Translation Quality} 
Figure \ref{fig:mt_models} presents averaged performance results for XLM-R Large with \flare on TyDiQA and NusaX, comparing the use of different-sized machine translation (MT) models, specifically NLLB 3.3B and NLLB 600M. The results demonstrate that \flare is robust to lower-quality machine translations. Although utilizing the larger NLLB 3.3B model yields performance improvements of 1.27\% and 1.77\% on NusaX and TyDiQA, respectively, the \flare models trained on lower-quality machine translations still achieve competitive performance with standard LoRA fine-tuning based on higher-quality machine translations. This demonstrates how \flare can further enhance resource efficiency by effectively leveraging smaller MT models, thereby reducing computational requirements without compromising performance relative to its benchmarks. 

\rparagraph{On Latent MT Fusion}
\begin{table}
\begin{center}
{
    \fontsize{8.0pt}{7.9pt}\selectfont
    \begin{tabular}{lcc}
    \toprule
    \bf Fusion Function & \bf TyDiQA & \bf NusaX \\
    \midrule
    \rowcolor{mintcream}
    \multicolumn{3}{p{7cm}}{\textit{Translate-Train (models are trained on data translated to the target language)}}\\
    \midrule
    add &  $40.76$ & $79.56$ \\
    mul &  $40.44$ & $78.81$ \\
    add+relu &  $40.93$ & $79.18$ \\
    cross-attention &  $39.63$ & $78.11$ \\
    \bottomrule
    \end{tabular}
}%
    \caption{Average performance of fusion functions using XLM-R Large with \texttt{FLARE}, evaluated on TyDiQA with Exact Match and on NusaX with Macro F1.}
    \label{tab:results_fusion_functions}
\end{center}
\end{table}

For encoder-only models like XLM-R Large and encoder-decoder models like mT5, latent MT fusion provides notable performance benefits compared to standard LoRA fine-tuning, X-Mixup, and \inputfusion, as shown in Table \ref{tab:results_fusion_functions}. However, for decoder-only models like Llama and Gemma, we do not observe substantial performance benefits from latent MT fusion. This suggests that reducing the computational resources for processing the source language representations in regular \flare can negatively impact cross-lingual transfer performance, particularly for larger models. Nonetheless, it provides a resource-efficient alternative to regular \flare for smaller mPLMs by avoiding the need for decoding in the MT and eliminating the forward pass for the source language representations.

\begin{table}
\begin{center}
\def\arraystretch{0.84}
{\fontsize{8.0pt}{7.9pt}\selectfont
    \begin{tabular}{lccc}
    \toprule
    \textbf{Model} & \textbf{\( r \)} & \textbf{TyDiQA} & \textbf{NusaX} \\
    \midrule
    \rowcolor{mintcream}
    \multicolumn{4}{p{7cm}}{\textit{Translate-Train (models are trained on training data translated to the target language)}} \\
    \midrule
    XLM-R Large w/ \flaremt & 8 & $40.86$ & $77.84$  \\
    w/ \flare &  & $42.37$ & $79.52$  \\
    \midrule
    XLM-R Large w/ \flaremt & 64 & $38.88$  & $77.18$ \\
    w/ \flare & & $40.93$  & $79.18$  \\
    \midrule
    XLM-R Large w/ \flaremt & 128 & $40.21$ & $77.18$ \\
    w/ \flare & & $40.88$ &  $78.32$ \\
    \bottomrule
    \end{tabular}
    }%
    \caption{Average performance for varying adapter bottleneck size \( r \) in LoRA; based on XLM-R Large, using \flare. Evaluation metrics include Exact Match for TyDiQA and Macro F1 for NusaX.}
    \label{tab:ablation_lora_r}
\end{center}
\end{table}

\rparagraph{Impact of Fusion Function} 
Our study on the impact of fusion functions, presented in Table~\ref{tab:results_fusion_functions}, shows that adding non-linearity to the fusion functions does not necessarily provide decisive performance benefits over simpler linear transformations. Notably, the functions \emph{add} and \emph{add+relu} demonstrate the best performance. Despite the additional parameters available in cross-attention, this technique does not yield superior downstream performance, consistent with the low performance of \xmixup in Table \ref{tab:results}. These findings suggest that the optimal fusion function is task-dependent and can be regarded as a hyperparameter that can be fine-tuned based on validation data.

\rparagraph{Impact of Adapter Capacity} 
We investigate the impact of the adapter capacity on \flare's performance, with results presented in Table \ref{tab:ablation_lora_r}. The findings indicate that small bottleneck sizes ($r=8$) yield optimal performance for XLM-R Large on the TyDiQA and NusaX datasets. This observation is consistent with the findings in the original LoRA paper \citep{hu2022lora}, indicating that the introduction of our fusion adapter does not affect the intrinsic rank of the tasks.

\begin{figure}
    \centering
    \includegraphics[width=\linewidth]{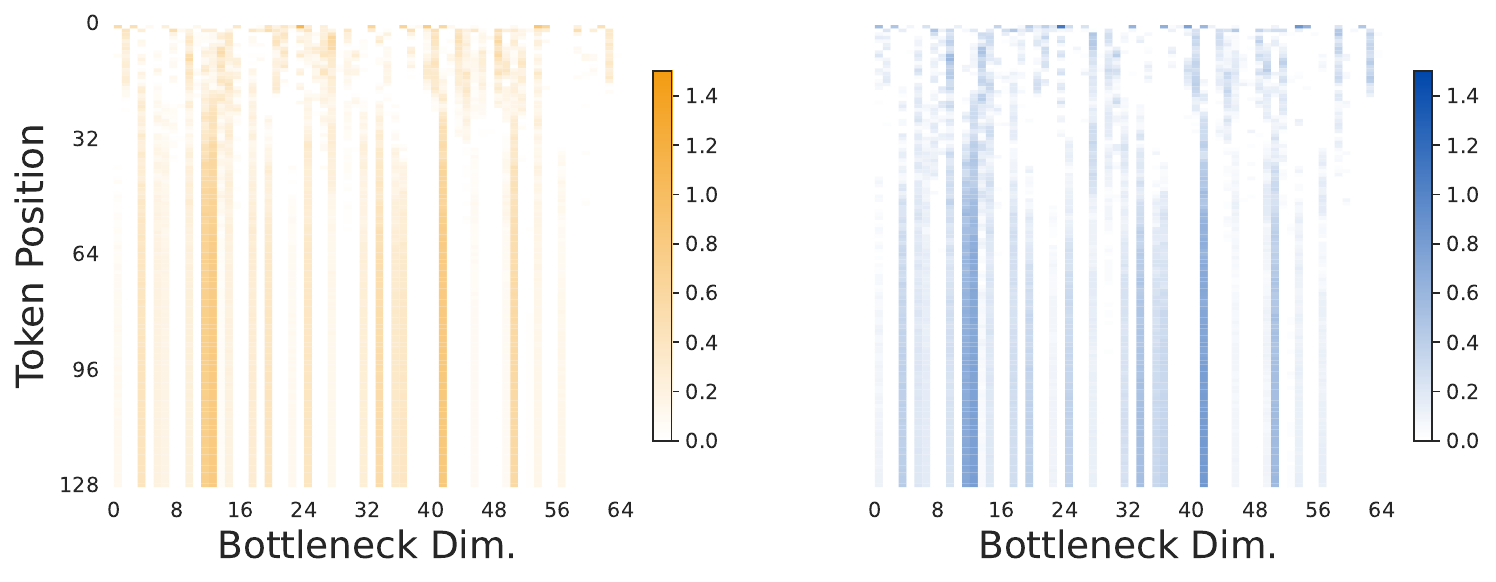}
    \caption{Average activation values for \textcolor{satorange}{English} and \textcolor{satblue}{Acehnese} in the first bottleneck query layer in XLM-R Large for the NusaX test set; \textit{add+relu} fusion.}
    \label{fig:activations}
    \vspace{-1mm}
\end{figure}

\begin{figure}
    \centering
    \includegraphics[width=\linewidth]{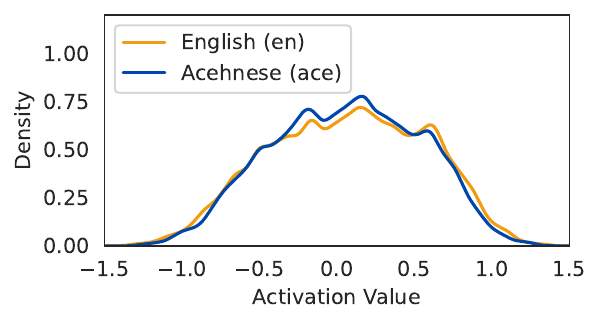}
    \caption{Average activations in the adapters across all XLM-R Large layers for the NusaX test set.}
    \label{fig:activations_mean}
    \vspace{-1mm}
\end{figure}

\rparagraph{Layer-wise Language Activation}
Figure~\ref{fig:activations_mean} shows that the magnitudes of source and target language activations across the entire XLM-R Large are comparable. This indicates that \flare does not overly rely on either source or target representations during fusion, but instead integrates both sources of information in a balanced manner. Further, Figure \ref{fig:activations} displays the average activations for English and Acehnese in the first adapter bottleneck: this confirms that both source and target languages maintain similar activation magnitudes. Hence, subsequent Acehnese representations are infused with the English representations from this initial transfer, integrating balanced source and target language information. Detailed activations for individual instances are illustrated in Figure \ref{fig:activations_instances}, which show positional activation differences and demonstrate the alignment of source and target languages for information transfer.

\section{Conclusion}

In this paper, we introduced Fusion for Language Representations (\flare), a parameter-efficient method for cross-lingual transfer (XLT) that enhances representation quality and downstream performance for languages other than English. Our experimental results demonstrate that \flare consistently outperforms strong XLT baselines, including target language fine-tuning with LoRA adapters, \xmixup, and \inputfusion, on various natural language understanding tasks. \flare demonstrates robust performance, even for lower-quality machine translations. A key takeaway is that \flare remains more parameter-efficient compared to benchmarked baseline approaches, while yielding superior performance. Furthermore, \flare provides most substantial performance benefits for multilingual question answering with decoder-only language models such as those from Llama and Gemma model families.

\section{Acknowledgments}
The resources and services used in this work were provided by the VSC (Flemish Supercomputer Center), funded by the Research Foundation - Flanders (FWO) and the Flemish Government. The work of Ivan Vulić has been supported by a personal Royal Society University Research Fellowship \textit{‘Inclusive and Sustainable Language Technology for a Truly Multilingual World’} (no 221137). Marie-Francine Moens was supported by the AIDAVA project (Horizon Europe: EU HORIZON-HLTH-2021-TOOL-06-03).

\section{Limitations}
Our work demonstrates that highly compressed English language representations can be effectively transferred to other languages within adapter bottlenecks. However, our experiments focus on bilingual transfer settings. Extending fusion adapters to integrate multiple target languages is non-trivial, as it requires adapters to extract language-agnostic information across multiple languages.

The proposed \flare method by design relies data availability for both source and target languages. Consequently, the application of \flare is dependent upon the availability of machine translation models.
Furthermore, our evaluation exclusively employs English as the high-resource source language for representation fusion. While English is predominantly used in mPLM pretraining corpora, exploring other high-resource languages that share linguistic similarities, with the target languages could potentially yield similar or improved cross-lingual transfer performance.

The reliance on machine translation also pose challenges for extremely low-resource languages, particularly where translation quality suffers or where cultural nuances are difficult to preserve \citep{alemayehu-etal-2024-error,conia-etal-2024-towards}. To examine the impact of translation quality on \flare, we conducted a manual error analysis on 200 randomly selected TyDiQA examples using Llama 3.1 and Gemma 2. We found no cases where machine translation quality was responsible for model errors. Most errors were due to answers unrelated to the question, while a smaller subset consisted of correct answers that did not match the dataset's ground truth. Additionally, the flexibility of \flare's fusion functions provides opportunities to incorporate cultural context more explicitly in future work.

We opt for the original and established LoRA architecture as our go-to PEFT model, due to its wide adoption and popularity. However, we note that \flare is not tied to the LoRA architecture, and might be combined with other, more recent and more sophisticated PEFT architectures~\cite{liu2024dora,kopiczko2024vera} in future research. 

Finally, our choice of base multilingual LMs has been motivated by the current state-of-the-art (SotA) in the field of multilingual NLP and XLT to low-resource languages for NLU tasks, combined with our computational budget and constraints. Therefore, the main models are SotA encoder-only (XLM-R) and encoder-decoder mPLMs (mT5), and decoder-only LLMs (Llama 3, Gemma 2). However, we note that the LLM technology and its adaptation to XLT for NLU in lower-resource languages has not been proven to be fully mature yet~\cite{lin2024mala,razumovskaia2024analyzing}.

% \section*{Acknowledgements}
% The resources and services used in this work were provided by the VSC (Flemish Supercomputer Center), funded by the Research Foundation - Flanders (FWO) and the Flemish Government.

\bibliography{anthology,custom}

\appendix

\section{Detailed Evaluation Results}

\begin{figure}
  \centering
  \subfloat{\includegraphics[width=\linewidth]{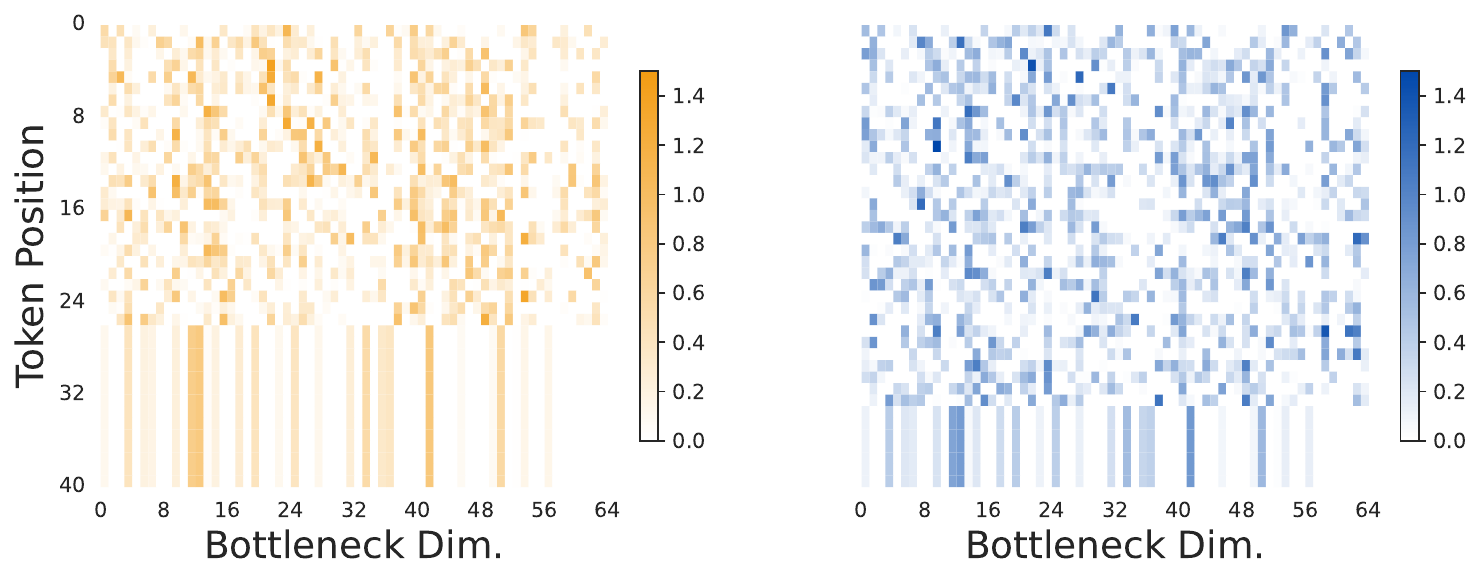}}
  \hfill
  \subfloat{\includegraphics[width=\linewidth]{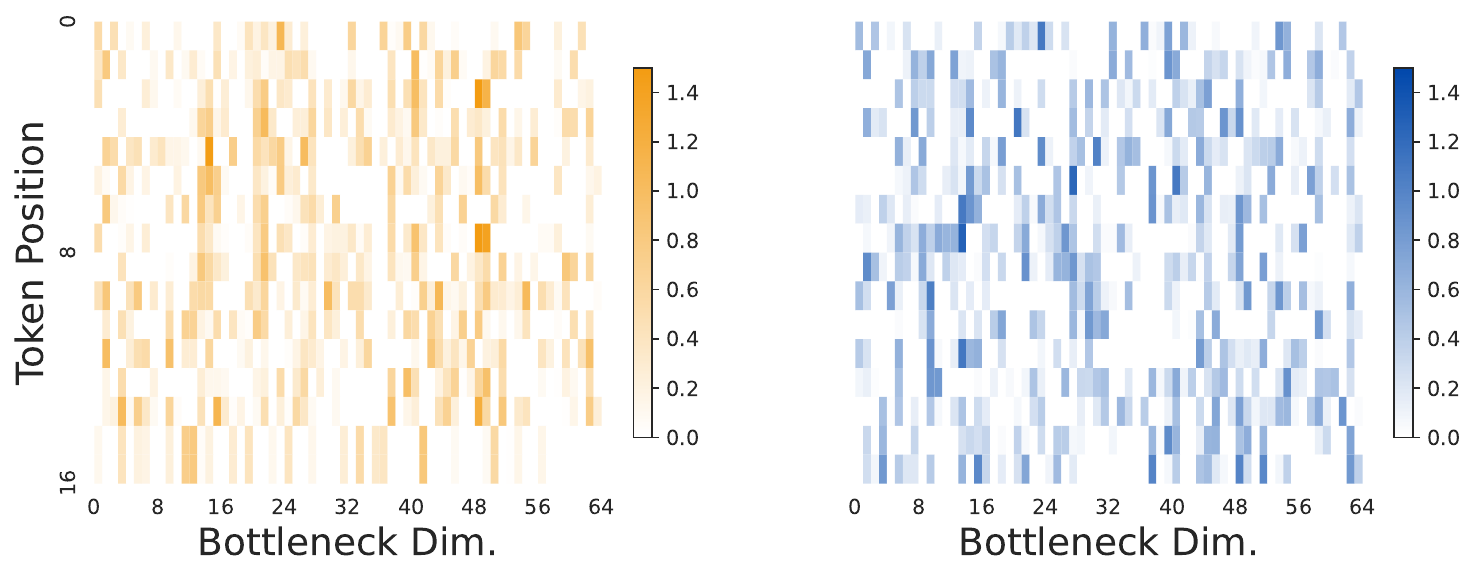}}
  \hfill
  \subfloat{\includegraphics[width=\linewidth]{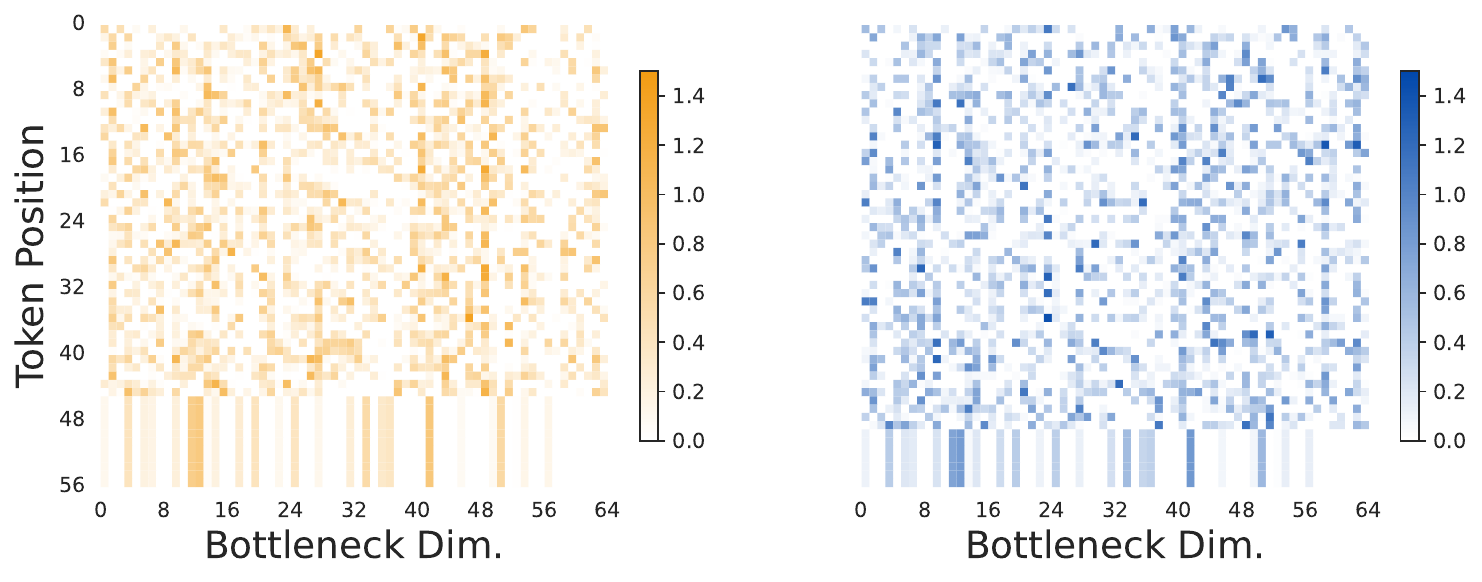}}
  
  \caption{Activation values for individual instances included in the NusaX test set. \textcolor{satorange}{English} and \textcolor{satblue}{Acehnese} activation values are extracted from the first bottleneck query layer in XLMR-Large, which is trained with the \textit{add+relu} fusion function.}
  \label{fig:activations_instances}
\end{figure}

\noindent \textbf{Figure \ref{fig:activations_instances}} displays average activations within the first adapter bottlenecks in the XLM-R Large model using \flare and the \emph{add+relu} fusion function. This visualization highlights the positional alignment process between English and Acehnese token representations, with varying activation values across different sequence positions reflecting the dynamics of language representation fusion.

\noindent \textbf{Table \ref{tab:ablation_ttrain_gold}} presents the performance of \flare and \inputfusion when using gold translations for fusion, as opposed to machine translations generated by NLLB. The results demonstrate that \inputfusion performance is sensitive to the quality of English input provided. Notably, when gold translations are available, \inputfusion replicates English performance, indicating that it heavily relies on the quality of English inputs. In contrast, \flare balances the fusion of source and target language information, as evident from the findings in Figure \ref{fig:activations_mean}. While \inputfusion outperforms \flare when gold translations are available, \flare achieves substantially higher performance in the more realistic setting using machine-translated data.

\noindent \textbf{Table \ref{tab:results_xnli_detail}} shows the results for the XNLI dataset for each language in zero-shot XLT, translate-test, translate-train settings, including translate-train with gold translations in the source language. The results confirm that \flare consistently improves XTL performance in the translate-train setting across different languages without particular bias towards typological relatedness to English or frequency in pretraining corpora.

\noindent \textbf{Table \ref{tab:results_tydiqa_detail}} details the results for the TyDiQA dataset for each language in the zero-shot XLT, translate-test, and translate-train settings. The outcomes demonstrate that \flare performance extends to tasks including positional information, such as extractive question-answering. 

\noindent \textbf{Table~\ref{tab:significance}} reports p-values from the Pitman permutation tests comparing \flare and \flaremt to baseline methods, based on average per-language performance from Tables~\ref{tab:results_xnli_detail}, \ref{tab:results_tydiqa_detail}, and \ref{tab:results_nusax_detail}. The results indicate that \flare and \flaremt yield consistent and often statistically significant improvements over regular LoRA, \xmixup, and \inputfusion, with particularly strong gains for the decoder-only Llama and Gemma models.

\noindent \textbf{Table \ref{tab:results_nusax_detail}} outlines the performance for the NusaX dataset for each language in zero-shot XLT, translate-test, translate-train, and translate-train settings with gold translations in the source language. Even with few training samples, our \flare method demonstrates consistent performance improvements across the low-resource languages included in the NusaX dataset.

\section{Training Details}
\label{sec:appendix_training_details}

Our evaluation results are averaged across \textit{five random seeds}. Initially, we fine-tune the language models on English task data using LoRA adapters set with $r=64$ and $\alpha=128$, which are subsequently integrated into the model's weights prior to task fine-tuning in the target languages. Hyperparameter configurations for each mPLM are provided in Table~\ref{tab:params}.

The total computation time for the experimental results exceeds 5,000 GPU hours. All models are trained using half-precision.

\section{Implementation Details of \flaremt}

We introduce \flaremt as variant of \flare aimed at further enhancing resource efficiency. \flaremt improves efficiency in two key ways. Firstly, it leverages latent translations generated by the machine translation (MT) encoder, thereby reducing the computational resources required to produce full text translations. Secondly, it eliminates the need for a forward pass through the source language representations in the mPLM, resulting in significant computational savings. As a results, a single source language representation, namely the latent translation, is fused with the target language representation in the fusion adapters for each transformer layer. To enable this fusion, a projection module is introduced to align the dimensions of the MT encoder with those of the mPLM. Although this module adds additional parameters, it is essential for ensuring compatibility between the two models. Notably, related work suggests that extending the single projection layer to a MLP and training it on additional self-supervised data can yield substantial performance benefits \citep{NEURIPS2023_6dcf277e,schmidt2024selfdistillation}. This provides a promising direction for future research and potential improvements to the \flaremt approach.

\section{Practical Implications}

The practical implementation of bilingual cross-lingual transfer methods, such as \flare, requires an additional step of language identification to determine bilingual adapter for model inference. While this introduces a preprocessing stage, language identification systems are widely accessible and highly accurate. For example, NLLB achieves a 95\% F1 score across 193 FLORES languages, including many low-resource languages \citep{burchell-etal-2023-open}, ensuring that this step can be seamlessly integrated into real-world applications.

\section{Another Ablation: Representation Fusion during Training Only}

To investigate the importance of utilizing source language representations during inference, we modified \flare to restrict representation fusion to the training phase only. Specifically, we limited the fusion with source language representations to 50\% of the training instances and excluded source language data during inference.
This evaluates cross-lingual transfer capabilities based on instance-independent patterns learned from source language representations during training. 
Our findings reveal that fusion adapters struggle to learn patterns that are independent of specific instances from source language representations during training. As a result, when implemented in the XLM-R Large model on the NusaX test set, the performance of the \emph{train-only} \flare variant decreased by 30\%. Crucially, this drop in performance underscores the importance of incorporating source language representations during inference to achieve effective cross-lingual adaptation.

\begin{table}
\begin{center}
\def\arraystretch{0.87}
{\fontsize{8.0pt}{7.9pt}\selectfont
    \begin{tabular}{lcc}
    \toprule
    \textbf{Model} & \textbf{XNLI} & \textbf{NusaX} \\
    \midrule
    \rowcolor{lightgold}
    \multicolumn{3}{p{7.2cm}}{\textit{Translate-Train (fusion models are trained on data translated into the target language and evaluated using \textbf{gold translations} from the target language to the source language)}}\\
    \midrule
    % XLM-R Base & & \\
    % w/ input-level fusion & $84.63$ & $87.87$ \\
    % w/ \texttt{FLARE} & $84.62$ & $75.43$ \\
    % \cmidrule{1-3}
    XLM-R Large & & \\
    w/ input-level fusion & $87.19$ & $90.93$  \\
    w/ \texttt{FLARE} & $88.15$ & $84.66$  \\
    \cmidrule{1-3}
    mT5-XL & & \\
    w/ input-level fusion & $89.67$ & $90.57$ \\
    w/ \texttt{FLARE} & $86.57$ & $80.72$ \\
    \bottomrule
    \end{tabular}
}%
    \caption{Average performance for the \emph{translate-train} setting with gold English translations during inference across languages included in the XNLI, and NusaX datasets, representing optimal translation quality. Evaluation metrics include accuracy for XNLI and Macro F1 for NusaX.}
    \label{tab:ablation_ttrain_gold}
\end{center}
\end{table}

\begin{table}
\begin{center}
\scalebox{0.7}{
    \begin{tabular}{llccc}
    \toprule
    \textbf{Model} & \textbf{Hparam} & \textbf{XNLI} & \textbf{TyDiQA} & \textbf{NusaX}\\
    \midrule
    % XLMR-Base & epochs & 10 \\
    %           & batch size & 32 \\
    %           & sequence length & \{128, 512, 128\} \\
    %           & learning rate & 2e-5 \\
    \addlinespace
    XLM-R Large & epochs & 10 & 10 & 20 \\
               & batch size & 64 & 64 & 64 \\
               & sequence length & 128 & 512 & 128 \\
               & learning rate & 2e-5 & 2e-4 & 2e-4 \\
    \addlinespace
    mT5-XL & epochs & 10 & 10 & 20 \\
           & batch size & 64 & 64 & 64 \\
           & sequence length & 128 & 512 & 128 \\
           & learning rate & 2e-5 & 2e-4 & 2e-4 \\
    \addlinespace
    Llama 3.1 8B & epochs & 3 & 3 & 5 \\
           & batch size & 64 & 64 & 64 \\
           & sequence length & 128 & 512 & 128 \\
           & learning rate & 2e-5 & 2e-4 & 2e-4 \\
    \addlinespace
    Gemma 2 9B & epochs & 3 & 3 & 5 \\
           & batch size & 64 & 64 & 64 \\
           & sequence length & 128 & 512 & 128 \\
           & learning rate & 2e-5 & 2e-4 & 2e-4 \\
    \bottomrule
    \end{tabular}}
    \caption{Hyperparameter configurations for each mPLM across the XNLI, TyDiQA, and NusaX datasets.}
    \label{tab:params}
\end{center}
\end{table}

\begin{table*}
\begin{center}
\scalebox{0.55}{
    % \begin{tabular}{lll>{\columncolor{gray!25}}c*{5}{c}|c}
    \begin{tabular}{l>{\columncolor{gray!25}}c*{14}{c}c}
    \toprule
    \textbf{Model} &
    % \textbf{Fusion Function} &
    \textbf{en} &
    \textbf{ar} &
    \textbf{bg} &
    \textbf{de} &
    \textbf{el} &
    \textbf{es} &
    \textbf{fr} &
    \textbf{hi} &
    \textbf{ru} &
    \textbf{sw} &
    \textbf{th} &
    \textbf{tr} &
    \textbf{ur} &
    \textbf{vi} &
    \textbf{zh} &
    \textbf{Avg.} \\
    \midrule

    \rowcolor{Gray}
    \multicolumn{17}{l}{\textit{Zero-Shot Cross-lingual Transfer}}\\
    \midrule
    % XLM-R Base  & $84.28$ & $71.61$ & $77.10$ & $75.76$ & $74.92$ & $78.27$ & $77.44$ & $69.20$ & $74.85$ & $63.11$ & $71.31$ & $71.27$ & $65.19$ & $74.06$ & $73.05$ & $72.65$ \\
    XLM-R Large & $87.81$ & $76.70$ & $81.37$ & $80.27$ & $80.24$ & $82.58$ & $81.56$ & $73.52$ & $78.31$ & $65.48$ & $76.01$ & $76.04$ & $69.43$ & $77.69$ & $78.07$ & $76.95$ \\
    mT5-XL & $89.04$ & $77.13$ & $82.27$ & $81.34$ & $81.00$ & $83.43$ & $82.74$ & $74.58$ & $80.22$ & $70.19$ & $75.93$ & $77.22$ & $70.35$ & $76.33$ & $78.15$ & $77.92$ \\
    Llama 3.1 8B & $91.47$ & $79.56$ & $80.56$ & $83.38$ & $80.73$ & $85.06$ & $84.07$ & $72.07$ & $81.66$ & $60.64$ & $75.38$ & $76.51$ & $62.64$ & $81.39$ & $80.01$ & $77.40$ \\
    Gemma 2 9B & $93.05$ & $80.38$ & $83.81$ & $84.42$ & $83.91$ & $85.05$ & $85.52$ & $77.59$ & $82.23$ & $73.96$ & $76.92$ & $79.27$ & $71.31$ & $81.54$ & $80.73$ & $\mathbf{80.47}$ \\
    
    \midrule
    \rowcolor{lavender}
    \multicolumn{17}{l}{\textit{Translate-Test (translate test data to English using NLLB 3.3B)}}\\
    \midrule
    % XLM-R Base & $ 84.28 $ & $ 74.42 $ & $ 77.64 $ & $ 78.35 $ & $ 77.90 $ & $ 80.05 $ & $ 78.44 $ & $ 72.56 $ & $ 75.80 $ & $ 70.29 $ & $ 70.65 $ & $ 75.92 $ & $ 66.38 $ & $ 75.64 $ & $ 72.94 $ & $ 74.78 $\\
    XLM-R Large & $87.81$ & $76.52$ & $81.46$ & $81.31$ & $81.03$ & $82.37$ & $81.57$ & $74.47$ & $77.80$ & $71.96$ & $72.65$ & $78.24$ & $67.87$ & $78.13$ & $74.43$ & $77.13$ \\
    mT5-XL & $89.04$ & $79.04$ & $83.15$ & $83.07$ & $82.56$ & $83.73$ & $83.30$ & $76.69$ & $80.47$ & $73.02$ & $74.69$ & $79.54$ & $69.80$ & $80.26$ & $77.15$ & $79.03$ \\
    Llama 3.1 8B & $91.47$ & $78.89$ & $83.61$ & $84.00$ & $82.86$ & $85.92$ & $83.97$ & $76.25$ & $80.00$ & $73.30$ & $73.34$ & $79.84$ & $69.03$ & $79.49$ & $76.55$ & $79.08$ \\
    Gemma 2 9B & $93.05$ & $79.76$ & $84.55$ & $84.99$ & $83.98$ & $87.04$ & $84.82$ & $77.01$ & $81.24$ & $73.75$ & $74.04$ & $81.31$ & $69.24$ & $80.57$ & $77.58$ & $\mathbf{79.99}$ \\

    \midrule
    \rowcolor{mintcream}
    \multicolumn{17}{l}{\textit{Translate-Train (models are trained on training data translated to the target language)}}\\
    \midrule
    
    XLM-R Large w/ LoRA & $87.81$ & $79.33$ & $84.13$ & $82.64$ & $83.21$ & $84.50$ & $83.06$ & $77.78$ & $81.42$ & $74.44$ & $79.73$ & $80.72$ & $74.71$ & $81.68$ & $79.44$ & $80.49$ \\
    w/ \xmixup & $87.81$ & $78.33$ & $82.48$ & $82.16$ & $80.12$ & $82.57$ & $81.23$ & $76.06$ & $80.54$ & $74.11$ & $79.48$ & $79.15$ & $73.67$ & $81.48$ & $81.18$ & $79.47$ \\
    w/ \inputfusion & $87.81$ & $77.29$ & $81.41$ & $81.07$ & $81.36$ & $82.40$ & $81.07$ & $74.89$ & $77.79$ & $72.13$ & $72.76$ & $78.47$ & $68.46$ & $78.11$ & $74.18$ & $77.24$ \\
    w/ \flaremt & $87.81$ & $80.91$ & $84.76$ & $84.12$ & $83.97$ & $85.03$ & $83.80$ & $79.04$ & $82.07$ & $76.71$ & $80.32$ & $81.70$ & $76.29$ & $81.88$ & $81.76$ & $\mathbf{81.60}$ \\
    w/ \flare & $87.81$ & $81.04$ & $84.12$ & $83.35$ & $83.44$ & $83.95$ & $83.53$ & $79.26$ & $79.58$ & $75.58$ & $80.40$ & $80.15$ & $75.50$ & $81.30$ & $82.71$ & $80.99$ \\
    \midrule
    
    mT5-XL w/ LoRA & $89.04$ & $79.44$ & $83.37$ & $83.23$ & $81.65$ & $84.17$ & $83.76$ & $76.84$ & $81.31$ & $75.97$ & $76.93$ & $77.68$ & $73.00$ & $79.13$ & $79.80$ & $79.73$ \\
    w/ \xmixup & $89.04$ & $80.14$ & $82.21$ & $82.37$ & $82.73$ & $82.87$ & $82.54$ & $77.16$ & $79.87$ & $76.10$ & $79.03$ & $78.00$ & $73.42$ & $79.97$ & $78.43$ & $79.63$ \\
    w/ \inputfusion & $89.04$ & $79.03$ & $82.76$ & $82.36$ & $82.14$ & $83.43$ & $82.89$ & $76.37$ & $80.28$ & $72.97$ & $75.11$ & $79.01$ & $69.73$ & $79.61$ & $77.70$ & $78.81$ \\
    w/ \flaremt & $89.04$ & $80.41$ & $83.73$ & $83.45$ & $82.91$ & $83.81$ & $83.57$ & $78.44$ & $81.35$ & $77.02$ & $78.62$ & $81.13$ & $75.46$ & $80.49$ & $80.75$ & $80.80$ \\
    w/ \flare & $89.04$ & $81.32$ & $83.72$ & $83.46$ & $82.23$ & $84.87$ & $83.47$ & $79.00$ & $81.28$ & $77.43$ & $79.54$ & $80.50$ & $74.27$ & $81.30$ & $81.66$ & $\mathbf{81.00}$ \\
    \midrule
    
    Llama 3.1 8B w/ LoRA & $91.47$ & $80.27$ & $82.88$ & $84.23$ & $83.16$ & $86.85$ & $85.49$ & $79.82$ & $83.97$ & $67.39$ & $77.58$ & $79.62$ & $76.94$ & $81.91$ & $80.37$ & $80.74$ \\
    w/ \xmixup & $91.47$ & $79.58$ & $82.94$ & $84.11$ & $81.61$ & $86.23$ & $85.39$ & $79.15$ & $83.16$ & $66.49$ & $76.51$ & $79.06$ & $77.20$ & $81.44$ & $80.18$ & $80.22$ \\
    w/ \inputfusion & $91.47$ & $79.17$ & $85.63$ & $85.39$ & $83.61$ & $87.10$ & $86.00$ & $77.55$ & $81.91$ & $74.72$ & $74.75$ & $82.32$ & $71.78$ & $82.07$ & $77.79$ & $80.70$ \\
    w/ \flaremt & $91.47$ & $80.27$ & $83.17$ & $84.87$ & $82.95$ & $86.75$ & $85.67$ & $80.35$ & $82.70$ & $67.19$ & $77.41$ & $79.95$ & $77.30$ & $81.92$ & $81.15$ & $80.83$ \\
    w/ \flare & $91.47$ & $80.00$ & $83.09$ & $84.92$ & $82.90$ & $86.55$ & $86.04$ & $80.80$ & $83.14$ & $67.08$ & $77.21$ & $79.33$ & $77.95$ & $82.33$ & $81.55$ & $\mathbf{80.92}$ \\
    \midrule
    
    Gemma 2 9B w/ LoRA & $93.05$ & $85.19$ & $87.87$ & $88.03$ & $87.78$ & $89.06$ & $87.13$ & $82.49$ & $86.10$ & $79.52$ & $83.20$ & $84.25$ & $78.07$ & $85.09$ & $84.69$ & $84.89$ \\
    w/ \xmixup & $93.05$ & $84.70$ & $87.76$ & $87.84$ & $87.32$ & $88.61$ & $87.83$ & $82.44$ & $85.27$ & $80.12$ & $82.60$ & $83.51$ & $77.35$ & $84.50$ & $84.86$ & $84.62$ \\
    w/ \inputfusion & $93.05$ & $79.98$ & $84.84$ & $85.19$ & $84.16$ & $86.65$ & $85.12$ & $77.39$ & $81.74$ & $74.48$ & $75.07$ & $81.98$ & $70.70$ & $81.74$ & $78.34$ & $80.53$ \\
    w/ \flaremt & $93.05$ & $85.07$ & $87.73$ & $87.89$ & $87.70$ & $88.93$ & $87.90$ & $82.69$ & $84.97$ & $80.52$ & $82.82$ & $84.18$ & $77.36$ & $84.88$ & $85.19$ & $84.84$ \\
    w/ \flare & $93.05$ & $84.67$ & $87.93$ & $88.14$ & $87.77$ & $89.23$ & $88.10$ & $82.86$ & $85.97$ & $79.73$ & $83.15$ & $84.08$ & $78.13$ & $85.23$ & $85.19$ & $\mathbf{85.01}$ \\
    
    \midrule
    \rowcolor{lightgold}
    \multicolumn{17}{l}{\textit{Translate-Train (fusion models are trained on data translated into the target language and evaluated using gold translations from the target language to the source language)}}\\
    \midrule

    XLM-R Large w/ \inputfusion & $ 87.81 $ & $ 88.41 $ & $ 88.54 $ & $ 88.46 $ & $ 88.36 $ & $ 88.28 $ & $ 88.02 $ & $ 88.38 $ & $ 85.91 $ & $ 86.23 $ & $ 85.91 $ & $ 85.85 $ & $ 86.05 $ & $ 85.85 $ & $ 86.45 $ & $ 87.19 $\\
    w/ \flare & $ 87.81 $ & $ 88.10 $ & $ 88.06 $ & $ 88.04 $ & $ 88.12 $ & $ 88.02 $ & $ 88.08 $ & $ 88.40 $ & $ 88.12 $ & $ 88.46 $ & $ 88.16 $ & $ 88.14 $ & $ 88.22 $ & $ 88.04 $ & $ 88.16 $ & $ \mathbf{88.15} $\\
    \midrule
    
    mT5-XL w/ \inputfusion & $ 89.04 $ & $ 90.04 $ & $ 89.80 $ & $ 89.54 $ & $ 89.70 $ & $ 89.78 $ & $ 89.50 $ & $ 89.80 $ & $ 89.52 $ & $ 89.56 $ & $ 89.84 $ & $ 89.66 $ & $ 89.38 $ & $ 89.52 $ & $ 89.70 $ & $ \mathbf{89.67}$ \\
    \flare & $ 89.04 $ & $ 88.62 $ & $ 88.74 $ & $ 88.80 $ & $ 85.34 $ & $ 87.83 $ & $ 86.19 $ & $ 84.31 $ & $ 86.12 $ & $ 89.66 $ & $ 88.49 $ & $ 89.56 $ & $ 79.22 $ & $ 85.33 $ & $ 83.73 $ & $ 86.57$ \\
    
    \bottomrule
    \end{tabular}}
    \caption{Average scores per language in the XNLI dataset. Model performance is evaluated using the Accuracy metric.}
    \label{tab:results_xnli_detail}
\end{center}
\end{table*}
\begin{table*}
\begin{center}
\scalebox{0.8}{
    % \begin{tabular}{lll>{\columncolor{gray!25}}c*{5}{c}|c}
    \begin{tabular}{l>{\columncolor{gray!25}}c*{8}{c}c}
    \toprule
    \textbf{Model} &
    \textbf{en} &
    \textbf{ar} &
    \textbf{ben} &
    \textbf{fi} &
    \textbf{ind} &
    \textbf{ko} &
    \textbf{ru} &
    \textbf{sw} &
    \textbf{tel} &
    \textbf{Avg.} \\
    \midrule

    \rowcolor{Gray}
    \multicolumn{11}{l}{\textit{Zero-Shot Cross-lingual Transfer}}\\
    \midrule
    XLM-R Large & $49.05$ & $24.27$ & $32.78$ & $30.15$ & $44.51$ & $29.92$ & $30.15$ & $40.27$ & $58.42$ & $36.31$ \\
    mT5-XL & $55.23$ & $30.94$ & $43.89$ & $35.56$ & $49.59$ & $41.59$ & $41.47$ & $50.63$ & $73.54$ & $\mathbf{45.90}$ \\
    Llama 3.1 8B & $55.61$ & $\phantom{0}2.41$ & $\phantom{0}0.00$ & $\phantom{0}5.22$ & $\phantom{0}1.64$ & $\phantom{0}0.58$ & $\phantom{0}5.49$ & $\phantom{0}2.52$ & $\phantom{0}1.06$ & $\phantom{0}2.36$ \\
    Gemma 2 9B & $60.08$ & $\phantom{0}1.63$ & $\phantom{0}2.46$ & $\phantom{0}3.20$ & $\phantom{0}0.88$ & $\phantom{0}0.24$ & $\phantom{0}2.77$ & $\phantom{0}1.87$ & $\phantom{0}6.61$ & $\phantom{0}2.46$ \\
    
    \midrule
    \rowcolor{lavender}
    \multicolumn{11}{l}{\textit{Translate-Test (translate test data to English using NLLB 3.3B)}}\\
    \midrule
    XLM-R Large & $49.05$ & $28.15$ & $52.78$ & $30.54$ & $47.79$ & $36.23$ & $34.92$ & $52.76$ & $45.30$ & $41.06$ \\
    mT5-XL  & $55.23$ & $34.01$ & $49.00$ & $36.08$ & $51.97$ & $42.61$ & $39.87$ & $54.95$ & $74.89$ & $\mathbf{47.92}$ \\
    Llama 3.1 8B & $55.61$ & $\phantom{0}1.35$ & $\phantom{0}5.00$ & $\phantom{0}2.21$ & $\phantom{0}0.99$ & $\phantom{0}0.58$ & $\phantom{0}2.71$ & $\phantom{0}4.68$ & $\phantom{0}2.75$ & $\phantom{0}2.53$ \\
    Gemma 2 9B & $60.08$ & $\phantom{0}0.68$ & $\phantom{0}3.33$ & $\phantom{0}1.74$ & $\phantom{0}0.66$ & $\phantom{0}0.00$ & $\phantom{0}1.68$ & $\phantom{0}1.08$ & $\phantom{0}9.04$ & $\phantom{0}2.28$ \\
    
    \midrule
    \rowcolor{mintcream}
    \multicolumn{11}{l}{\textit{Translate-Train (models are trained on training data translated to the target language)}}\\
    \midrule

    XLM-R Large w/ LoRA & $49.05$ & $31.10$ & $38.97$ & $30.35$ & $46.61$ & $37.11$ & $24.54$ & $47.13$ & $65.34$ & $40.14$ \\
    w/ \xmixup & $49.05$ & $26.25$ & $36.67$ & $26.65$ & $43.70$ & $34.19$ & $24.40$ & $45.85$ & $68.21$ & $38.24$ \\
    w/ \inputfusion & $49.05$ & $31.56$ & $40.00$ & $30.04$ & $45.63$ & $33.33$ & $28.12$ & $50.29$ & $64.62$ & $40.45$ \\
    w/ \flaremt & $49.05$ & $30.21$ & $35.00$ & $33.02$ & $44.82$ & $35.90$ & $25.38$ & $48.46$ & $58.26$ & $38.88$ \\
    w/ \flare & $49.05$ & $30.83$ & $41.67$ & $33.44$ & $44.61$ & $35.33$ & $26.17$ & $48.47$ & $66.93$ & $\mathbf{40.93}$ \\
    \midrule
    
    mT5-XL w/ LoRA & $55.23$ & $33.48$ & $46.71$ & $38.90$ & $49.84$ & $46.77$ & $33.20$ & $51.46$ & $73.73$ & $46.76$ \\
    w/ \xmixup & $55.23$ & $32.55$ & $54.49$ & $38.15$ & $52.17$ & $49.05$ & $33.75$ & $52.04$ & $73.68$ & $48.23$ \\
    w/ \inputfusion & $55.23$ & $34.75$ & $49.74$ & $39.29$ & $51.74$ & $45.29$ & $30.69$ & $52.79$ & $76.32$ & $47.58$ \\
    w/ \flaremt & $55.23$ & $46.08$ & $48.59$ & $39.31$ & $53.94$ & $44.90$ & $30.14$ & $49.16$ & $75.75$ & $48.48$ \\
    w/ \flare & $55.23$ & $47.86$ & $49.55$ & $40.98$ & $54.23$ & $46.05$ & $30.41$ & $50.85$ & $74.82$ & $\mathbf{49.34}$ \\
    \midrule
    
    Llama 3.1 8B w/ LoRA & $55.61$ & $39.69$ & $26.11$ & $44.27$ & $56.61$ & $53.56$ & $37.23$ & $53.47$ & $31.75$ & $42.84$ \\
    w/ \xmixup & $55.61$ & $23.44$ & $16.11$ & $37.26$ & $30.69$ & $0.00$ & $10.96$ & $0.00$ & $21.32$ & $17.47$ \\
    w/ \inputfusion & $55.61$ & $37.48$ & $21.03$ & $45.86$ & $56.68$ & $62.61$ & $37.03$ & $61.82$ & $46.20$ & $46.09$ \\
    w/ \flaremt & $55.61$ & $38.33$ & $26.11$ & $37.90$ & $48.78$ & $53.85$ & $34.04$ & $44.94$ & $27.63$ & $38.95$ \\
    w/ \flare & $55.61$ & $44.48$ & $26.66$ & $48.09$ & $62.80$ & $56.98$ & $42.44$ & $59.73$ & $40.70$ & $\mathbf{47.74}$ \\
    \midrule
    
    Gemma 2 9B w/ LoRA & $60.08$ & $43.75$ & $46.67$ & $44.69$ & $57.52$ & $59.83$ & $38.82$ & $59.72$ & $48.42$ & $49.93$ \\
    w/ \xmixup & $60.08$ & $37.71$ & $20.00$ & $46.92$ & $54.07$ & $39.32$ & $14.33$ & $25.71$ & $45.54$ & $35.45$ \\
    w/ \inputfusion & $60.08$ & $45.74$ & $50.14$ & $40.45$ & $59.97$ & $61.87$ & $41.06$ & $61.91$ & $49.14$ & $\mathbf{51.29}$ \\
    w/ \flaremt & $60.08$ & $43.23$ & $45.00$ & $44.05$ & $59.45$ & $57.83$ & $40.05$ & $58.82$ & $48.60$ & $49.63$ \\
    w/ \flare & $60.08$ & $44.79$ & $46.67$ & $47.35$ & $65.24$ & $60.68$ & $41.82$ & $60.75$ & $49.83$ & $52.14$ \\
    
    \bottomrule
    \end{tabular}}
    \caption{Average scores per language in the TyDiQA dataset. Model performance is evaluated using the Exact Match metrics.}
    \label{tab:results_tydiqa_detail}
\end{center}
\end{table*}
\begin{table*}
\begin{center}
\scalebox{0.55}{
    \begin{tabular}{l>{\columncolor{gray!25}}c*{10}{c}c}
    \toprule
    \textbf{Model} &
    \textbf{en} &
    \textbf{ace} &
    \textbf{ban} &
    \textbf{bjn} &
    \textbf{bug} &
    \textbf{ind} &
    \textbf{jav} &
    \textbf{mad} &
    \textbf{min} &
    \textbf{nij} &
    \textbf{sun} &
    \textbf{Avg.} \\
    \midrule

    \rowcolor{Gray}
    \multicolumn{13}{l}{\textit{Zero-Shot Cross-lingual Transfer}}\\
    \midrule
    XLM-R Large & $92.04$ & $68.34$ & $75.37$ & $80.37$ & $51.90$ & $90.76$ & $84.69$ & $69.01$ & $80.06$ & $69.23$ & $82.89$ & $\mathbf{75.26}$ \\
    mT5-XL & $91.77$ & $72.26$ & $76.42$ & $79.79$ & $49.51$ & $90.61$ & $87.49$ & $61.38$ & $77.71$ & $65.31$ & $86.73$ & $74.72$ \\
    Llama 3.1 8B &  $89.75$ & $70.50$ & $72.00$ & $80.33$ & $39.92$ & $89.75$ & $77.25$ & $64.75$ & $77.75$ & $65.42$ & $79.75$ & $71.74$ \\
    Gemma 2 9B & $91.15$ & $66.42$ & $71.58$ & $82.08$ & $31.92$ & $91.67$ & $86.25$ & $64.00$ & $80.75$ & $64.33$ & $77.08$ & $71.61$ \\
    
    \midrule
    \rowcolor{lavender}
    \multicolumn{13}{l}{\textit{Translate-Test (translate test data to English using NLLB 3.3B)}}\\
    \midrule
    XLM-R Large & $92.04$ & $73.20$ & $73.88$ & $82.09$ & $60.47$ & $88.85$ & $84.27$ & $61.24$ & $81.19$ & $59.35$ & $83.97$ & $74.85$ \\
    mT5-XL & $91.77$ & $76.27$ & $73.43$ & $81.72$ & $69.29$ & $86.86$ & $83.50$ & $60.63$ & $82.47$ & $60.86$ & $82.68$ & $\mathbf{75.77}$ \\
    Llama 3.1 8B & $89.75$ & $70.83$ & $73.00$ & $80.75$ & $39.92$ & $89.58$ & $78.00$ & $65.25$ & $81.58$ & $67.33$ & $80.42$ & $72.67$ \\
    Gemma 2 9B & $91.15$ & $66.42$ & $71.58$ & $82.08$ & $31.92$ & $91.67$ & $86.25$ & $64.00$ & $80.75$ & $64.33$ & $77.08$ & $71.61$ \\
    
    \midrule
    \rowcolor{mintcream}
    \multicolumn{13}{l}{\textit{Translate-Train (models are trained on training data translated to the target language)}}\\
    \midrule
    
    XLM-R Large w/ LoRA & $92.04$ & $74.19$ & $74.55$ & $81.84$ & $60.99$ & $89.40$ & $85.90$ & $70.75$ & $81.15$ & $67.35$ & $83.87$ & $77.00$ \\
    w/ \xmixup & $92.04$ & $73.10$ & $73.08$ & $81.18$ & $62.22$ & $88.38$ & $85.90$ & $65.79$ & $82.30$ & $68.97$ & $82.74$ & $76.37$ \\
    \inputfusion & $92.04$ & $77.77$ & $75.89$ & $82.67$ & $69.96$ & $89.44$ & $87.92$ & $66.66$ & $79.55$ & $68.47$ & $87.01$ & $78.53$ \\
    w/ \flaremt & $92.04$ & $73.33$ & $75.95$ & $81.13$ & $57.20$ & $90.76$ & $86.59$ & $69.77$ & $83.42$ & $68.90$ & $84.73$ & $77.18$ \\
    w/ \flare & $92.04$ & $76.47$ & $77.27$ & $80.71$ & $70.18$ & $90.54$ & $87.42$ & $71.33$ & $85.15$ & $70.16$ & $82.59$ & $\mathbf{79.18}$ \\
    \midrule
    
    mT5-XL w/ LoRA & $91.77$ & $80.66$ & $81.92$ & $85.83$ & $65.36$ & $89.78$ & $90.40$ & $69.85$ & $82.30$ & $69.27$ & $88.76$ & $80.41$ \\
    w/ \xmixup & $91.77$ & $80.34$ & $74.60$ & $83.76$ & $68.87$ & $88.52$ & $88.75$ & $68.25$ & $83.66$ & $65.60$ & $83.76$ & $78.61$ \\
    \inputfusion & $91.77$ & $81.00$ & $79.48$ & $85.54$ & $71.44$ & $89.75$ & $87.58$ & $66.33$ & $83.28$ & $68.02$ & $88.78$ & $80.12$ \\
    w/ \flaremt & $91.77$ & $81.19$ & $84.12$ & $85.19$ & $66.59$ & $90.14$ & $89.67$ & $71.16$ & $84.80$ & $71.87$ & $88.94$ & $\mathbf{81.37}$ \\
    w/ \flare & $91.77$ & $81.03$ & $82.03$ & $85.88$ & $66.95$ & $89.55$ & $89.80$ & $68.63$ & $84.20$ & $69.31$ & $88.05$ & $80.54$ \\
    \midrule
    
    Llama 3.1 8B w/ LoRA & $89.75$ & $76.26$ & $73.71$ & $78.10$ & $62.82$ & $88.66$ & $84.29$ & $62.91$ & $82.20$ & $58.04$ & $80.64$ & $74.76$ \\
    w/ \xmixup & $89.75$ & $77.25$ & $76.58$ & $79.00$ & $64.17$ & $89.92$ & $85.08$ & $64.58$ & $82.17$ & $59.83$ & $80.50$ & $75.91$ \\
    \inputfusion & $89.75$ & $74.83$ & $66.17$ & $80.17$ & $66.67$ & $89.17$ & $85.50$ & $59.63$ & $82.63$ & $57.25$ & $84.00$ & $74.60$ \\
    w/ \flaremt& $89.75$ & $78.21$ & $72.00$ & $74.29$ & $64.21$ & $87.96$ & $83.04$ & $64.38$ & $80.75$ & $62.38$ & $77.96$ & $74.52$ \\
    w/ \flare & $89.75$ & $78.88$ & $75.25$ & $80.25$ & $64.25$ & $91.17$ & $85.88$ & $65.13$ & $81.38$ & $57.88$ & $80.75$ & $\mathbf{76.08}$ \\
    \midrule
    
    Gemma 2 9B w/ LoRA & $91.15$ & $77.89$ & $79.14$ & $82.30$ & $66.83$ & $91.71$ & $87.56$ & $69.81$ & $85.72$ & $67.61$ & $85.18$ & $79.37$ \\
    w/ \xmixup & $91.15$ & $80.94$ & $77.92$ & $82.82$ & $68.22$ & $91.46$ & $88.29$ & $69.41$ & $87.64$ & $67.26$ & $85.48$ & $79.94$ \\
    \inputfusion & $91.15$ & $77.67$ & $76.88$ & $83.50$ & $70.58$ & $89.92$ & $87.17$ & $63.92$ & $84.50$ & $60.71$ & $84.92$ & $77.98$ \\
    w/ \flaremt & $91.15$ & $79.88$ & $80.67$ & $81.88$ & $62.50$ & $91.04$ & $85.67$ & $66.04$ & $84.71$ & $64.42$ & $84.08$ & $78.09$ \\
    w/ \flare & $91.15$ & $82.75$ & $81.00$ & $83.17$ & $65.08$ & $92.83$ & $86.83$ & $73.08$ & $87.75$ & $70.58$ & $85.50$ & $\mathbf{80.86}$ \\
    
    \midrule
    \rowcolor{lightgold}
    \multicolumn{13}{l}{\textit{Translate-Train (fusion models are trained on data translated into the target language and evaluated using gold translations from the target language to the source language)}}\\
    \midrule
   
    XLM-R Large \inputfusion & $92.04$ & $91.24$ & $91.08$ & $90.55$ & $90.69$ & $91.99$ & $90.88$ & $91.23$ & $91.07$ & $90.07$ & $90.52$ & $\mathbf{90.93}$ \\
    w/ \flare & $92.04$  & $89.24$ & $88.98$ & $82.55$ & $90.07$ & $90.22$ & $88.15$ & $71.20$ & $87.58$ & $72.93$ & $85.71$ & $84.66$ \\
    \midrule

    mT5-XL \inputfusion & $91.77$ & $91.39$ & $90.39$ & $91.47$ & $91.54$ & $90.88$ & $89.49$ & $88.87$ & $90.86$ & $89.20$ & $91.60$ & $\mathbf{90.57}$ \\
    w/ \flare & $91.77$  & $83.80$ & $80.55$ & $84.06$ & $64.70$ & $88.32$ & $90.50$ & $74.36$ & $83.64$ & $69.29$ & $88.00$ & $80.72$ \\
    
    \bottomrule
    \end{tabular}}
    \caption{Average scores per language in the NusaX dataset. Model performance is evaluated using the Macro F1 metric.}
    \label{tab:results_nusax_detail}
\end{center}
\end{table*}
\begin{table*}
\begin{center}
\vspace{-0.5mm}
{
\def\arraystretch{0.99}
\fontsize{8.0pt}{7.9pt}\selectfont
%\begin{tabular}{@{\extracolsep{\fill}} lcccc}
\begin{tabularx}{\textwidth}{ll YYY}
\toprule
\textbf{Dataset} & \textbf{Model} & \textbf{LoRA} & \textbf{\xmixup} & \textbf{Input-level fusion} \\
\midrule
\rowcolor{mintcream}
\multicolumn{5}{l}{\textit{\textbf{Translate-Train} (models are trained on training data translated to the target language)}} \\
\midrule
XNLI   & XLM-R w/ \flaremt & $0.000$***                  & $0.000$***                    & $0.000$***                          \\
       & w/ \flare   & $0.139$\phantom{***}                  & $0.001$***                    & $0.001$***                          \\
\cmidrule{2-5}
       & mT5-XL w/ \flaremt & $0.001$***                  & $0.000$***                    & $0.000$***                          \\
       & w/ \flare   & $0.001$***                  & $0.000$***                    & $0.000$***                          \\
\cmidrule{2-5}
       & Llama 3.1 8B w/ \flaremt & $0.564$\phantom{***}                  & $0.001$***                    & $0.885$\phantom{***}                          \\
       & w/ \flare   & $0.321$\phantom{***}                  & $0.000$***                    & $0.813$\phantom{***}                          \\
\cmidrule{2-5}
       & Gemma 2 9B w/ \flaremt & $0.764$\phantom{***}                  & $0.006$***                    & $0.000$***                          \\
       & w/ \flare   & $0.221$\phantom{***}                  & $0.002$***                    & $0.000$***                          \\
\midrule
TyDiQA & XLM-R w/ \flaremt & $0.311$\phantom{***}                  & $0.710$\phantom{***}                    & $0.213$\phantom{***}                          \\
       & w/ \flare   & $0.296$\phantom{***}                  & $0.035$**\phantom{*}                    & $0.531$\phantom{***}                          \\
\cmidrule{2-5}
       & mT5-XL w/ \flaremt & $0.434$\phantom{***}                  & $0.938$\phantom{***}                    & $0.772$\phantom{***}                          \\
       & w/ \flare   & $0.181$\phantom{***}                  & $0.804$\phantom{***}                    & $0.393$\phantom{***}                          \\
\cmidrule{2-5}
       & Llama 3.1 8B w/ \flaremt & $0.029$**\phantom{*}                  & $0.007$***                    & $0.005$***                          \\
       & w/ \flare   & $0.007$***                  & $0.007$***                    & $0.378$\phantom{***}                          \\
\cmidrule{2-5}
       & Gemma 2 9B w/ \flaremt & $0.536$\phantom{***}                  & $0.014$**\phantom{*}                    & $0.113$\phantom{***}                          \\
       & w/ \flare   & $0.015$**\phantom{*}                  & $0.007$***                    & $0.505$\phantom{***}                          \\
\midrule
NusaX  & XLM-R  w/ \flaremt & $0.773$\phantom{***}                  & $0.323$\phantom{***}                    & $0.462$\phantom{***}                          \\
       & w/ \flare   & $0.026$**\phantom{*}                  & $0.009$***                    & $0.514$\phantom{***}                          \\
\cmidrule{2-5}
       & mT5-XL w/ \flaremt & $0.044$**\phantom{*}                  & $0.023$**\phantom{*}                    & $0.192$\phantom{***}                          \\
       & w/ \flare   & $0.693$\phantom{***}                  & $0.032$**\phantom{*}                    & $0.543$\phantom{***}                          \\
\cmidrule{2-5}
       & Llama 3.1 8B w/ \flaremt & $0.750$\phantom{***}                  & $0.094$*\phantom{**}                    & $0.964$\phantom{***}                          \\
       & w/ \flare   & $0.015$**\phantom{*}                  & $0.632$\phantom{***}                    & $0.265$\phantom{***}                          \\
\cmidrule{2-5}
       & Gemma 2 9B w/ \flaremt & $0.086$*\phantom{**}                  & $0.027$**\phantom{*}                    & $0.936$\phantom{***}                          \\
       & w/ \flare   & $0.041$**\phantom{*}                  & $0.187$\phantom{***}                    & $0.088$*\phantom{**} \\                 
\bottomrule
\end{tabularx}
}
\caption{P-values from the Pitman permutation test comparing \flare and \flaremt (rows) against baseline methods (columns) on the XNLI, TyDiQA, and NusaX datasets. The test is performed on average performance scores per language in the translate-train setting, as reported in Tables~\ref{tab:results_xnli_detail}, \ref{tab:results_tydiqa_detail}, and \ref{tab:results_nusax_detail}. Statistical significance at the 90\%, 95\%, and 99\% confidence levels is indicated by *, **, and ***, respectively.}

\label{tab:significance}
\end{center}
\end{table*}
\begin{table*}
\begin{center}
{
\def\arraystretch{1.05}
\fontsize{9pt}{9pt}\selectfont
\scalebox{1}{\begin{tabularx}{\textwidth}{l lll}
\toprule
\textbf{Task} &
\textbf{Language} &
\textbf{ISO Code} &
\textbf{Source} \\
\midrule

XNLI & Arabic & ar & \multirow{14}{*}{Crowd-sourced \citep{williams-etal-2018-broad}} \\
& Bulgarian & bg & \\
& Chinese & zh & \\
& French & fr & \\
& German & de & \\
& Greek & el & \\
& Hindi & hi & \\
& Russian & ru & \\
& Spanish & es & \\
& Swahili & sw & \\
& Thai & th & \\
& Turkish & tr & \\
& Urdu & ur & \\
& Vietnamese & vi & \\

\midrule 

TyDiQA & Arabic & ar & \multirow{8}{*}{Wikipedia \citep{clark-etal-2020-tydi}}\\
& Bengali & ben & \\
& Finnish & fi & \\
& Indonesian & ind & \\
& Korean & ko & \\
& Russian & ru & \\
& Swahili & sw & \\
& Telugu & tel & \\

\midrule 

NusaX & Acehnese & ace & \multirow{9}{*}{SmSA \citep{Purwarianti_smsa}}\\
& Balinese & ban & \\
& Banjarese & bjn & \\
& Buginese & bug & \\
& Indonesian & ind & \\
& Javanese & jav & \\
& Madurese & mad & \\
& Minangkabau & min & \\
& Ngaju & nij & \\

\bottomrule
\end{tabularx}}
}
\caption{Overview of languages and corresponding source data used in the experiments, categorized by task.}
\label{tab:tasks_langs}
\end{center}
\end{table*}

\end{document}